\documentclass[11pt]{article}

\usepackage[final]{acl}
\usepackage{siunitx}
\usepackage{amsmath}
\usepackage[dvipsnames]{xcolor}
\usepackage{times}
\usepackage{latexsym}
\usepackage{listings}
\usepackage[T1]{fontenc}

\usepackage[utf8]{inputenc}

\usepackage{microtype}

\usepackage{inconsolata}

\usepackage{graphicx}
\usepackage{dblfloatfix} 
\usepackage{svg}
\usepackage{booktabs}
\usepackage{hyperref}

\usepackage{soul}
\usepackage{tcolorbox}
\usepackage{enumitem}
\usepackage{titlesec}
\usepackage{makecell}   

\usepackage{xcolor}
\usepackage{colortbl}

\definecolor{PolishColor}{HTML}{4C72B0}   %
\definecolor{GenColor}{HTML}{55A868}      %
\definecolor{MedColor}{HTML}{C44E52}      %

\definecolor{PolishBG}{HTML}{EAF1FB}
\definecolor{GenBG}{HTML}{EAF6ED}
\definecolor{MedBG}{HTML}{FAEAEA}

\tcbset{
    questionbox/.style={
        colback=gray!5,
        colframe=black!50,
        boxrule=0.5pt,
        arc=2mm,
        left=2mm,
        right=2mm,
        top=1mm,
        bottom=1mm
    }
}

\title{Reassessing High-Performing LLMs on Polish Medical Exams: True Competence or Bias-Driven Performance?}

\author{
Antoni Lasik$^{1}$~~~%
Jakub Pokrywka$^{2}$~~~%
Łukasz Grzybowski$^{2,3}$~~~%
Jeremi Ignacy Kaczmarek$^{2,4,6}$~~~%
\AND
\vspace{-4em}
Gabriela Korzańska$^{4}$~~~%
Janusz Świeczkowski-Feiz$^{5,7}$~~~%
Oskar Pastuszek$^{6}$~~~%
\AND
Paulina Hoffman$^{7}$~~~%
Jakub Tomasz Dąbrowski$^{5}$~~~%
Wojciech Kusa$^{1}$~~~%
\vspace{1em}
\\
$^{1}$NASK National Research Institute~~~%
$^{2}$Adam Mickiewicz University~~~%
$^{3}$ARAAI Poland~~~%
\vspace{0.2em}
\\
$^{4}$Poznań University of Medical Sciences~~~%
$^{5}$Centre of Postgraduate Medical Education, Poland~~~%
\vspace{0.2em}
\\
$^{6}$T. Marciniak Lower Silesian Specialist Hospital~~~%
$^{7}$Medical University of Warsaw~~~%
\vspace{0.5em}
\\
 \small{
   \textbf{Correspondence:} \texttt{\{firstname.lastname\}@nask.pl} %
 }
}

\begin{document}
\maketitle
\begin{abstract}
Large language models (LLMs) in medicine are mainly evaluated using multiple-choice question answering (MCQA), which can overestimate real clinical ability due to guessing strategies and answer biases. 
To address these limitations, we introduce an expanded and more challenging benchmark based on Polish medical exams, adding over 15,000 questions, two new domains, and four structural modifications that reduce MCQA-specific artifacts and better test reasoning.
We evaluate 21 LLMs and show that evaluation design strongly affects results. Under our harder setup, the best model (Qwen3.5-122B) drops by 28.4 and 31 pp on English and Polish exams, respectively. Despite low evidence of data contamination, standard MCQA scores do not reliably reflect true medical competence.
To facilitate further research, we make our benchmark publicly available.\footnote{\url{https://github.com/NASK-NLP/Polish-Medical-Exams}}
\end{abstract}

\section{Introduction}

Large language models (LLMs) are increasingly explored for medical applications \cite{yang2023large}, motivating the need for reliable and informative evaluation frameworks. In recent years, a wide range of benchmarks has been introduced to track progress in medical LLM capabilities, with most evaluations relying on multiple-choice questions answering (MCQA). 

However, MCQA can overestimate performance and fail to capture reasoning depth, uncertainty handling due to the constrained answer space, where models may exploit statistical regularities, answer priors, and inter-choice relationships rather than perform genuine problem-solving. The presence of explicit options enables elimination strategies, pattern matching, and reconstruction of plausible questions from the choices themselves, which can inflate accuracy without corresponding improvements in underlying reasoning capabilities \cite{balepur-etal-2024-artifacts, balepur-etal-2025-best}.  This is particularly important in the medical domain, where the goal is not only to compare LLMs against one another, but also to accurately assess the limitations of their knowledge, as inadequately evaluated models may lead to harmful consequences in healthcare.

In addition, LLM performance in the medical domain varies appreciably across languages \cite{alonso2024medexpqa, jin2024better}. These findings suggest that medical questions require not only general biomedical knowledge but also local contextual understanding, including region-specific disease patterns, health-system constraints, and socio-cultural factors \cite{nimo2025afrimed}. This highlights the need for evaluation benchmarks grounded in local medical practice and linguistic context.

\begin{figure*}[th]
    \centering
    \includegraphics[width=\linewidth]{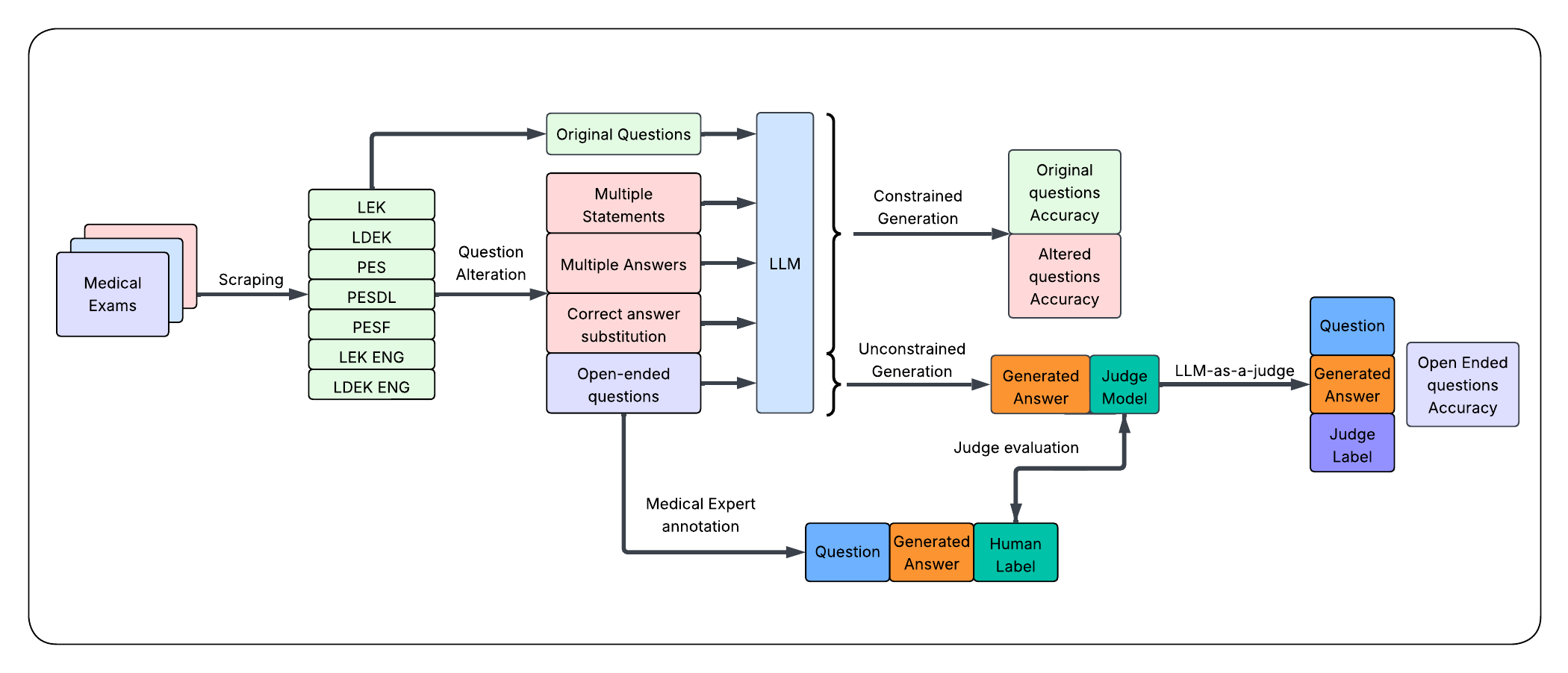}
    \caption{Overview of our methodology.}
    \label{fig:overview}
\end{figure*}

Considering the limitations inherent to MCQA benchmarks, we extend and refine the previously introduced medical benchmark for Polish medical examinations \cite{grzybowski2025polish}. The objective of this work is to develop a more rigorous evaluation framework that more accurately reflects the underlying knowledge of LLMs by mitigating their ability to exploit structural properties of the MCQA format. Furthermore, we systematically assess the discrepancies between conventional MCQA-based performance and results obtained under the proposed evaluation protocol.

Our contributions are as follows:
\begin{itemize}
\item We extend the previously introduced Polish medical knowledge evaluation dataset\footnote{\url{https://huggingface.co/spaces/amu-cai/Polish_Medical_Exams}}
 by incorporating over 15,000 additional questions from the Pharmaceutical Specialist Examination and the Examination for Laboratory Diagnosticians Specialization, and providing updates for other exams.
\item We introduce structural alterations to the question format and an open-ended setup, resulting in a refined evaluation protocol supported by a recomposed and more challenging benchmark, designed to better assess true LLM knowledge while limiting the exploitation of MCQA-specific artifacts.
\item We conduct a comprehensive evaluation of 21 LLMs spanning diverse model families, including general-purpose models, domain-specific medical models, and Polish-language models, enabling a systematic comparison of their performance on MCQA tasks and their underlying medical knowledge.
\end{itemize}

\section{Related work} \label{sec:related}

\subsection{MCQA evaluation}

Authors of \cite{balepur-etal-2024-artifacts} showed that LLMs can achieve above-baseline accuracy in multiple-choice settings even when only answer options are provided, with gains observed in 11 out of 12 evaluated cases and improvements of up to 0.33 accuracy. Their analysis indicates that this behavior cannot be explained solely by memorization or simple priors over individual answers; instead, models exploit group dynamics among choices and are sometimes able to infer plausible underlying questions from the options alone. In a follow-up study, \cite{balepur-etal-2025-best} identify fundamental limitations of MCQA as an evaluation paradigm, showing that it poorly captures generative reasoning, is susceptible to shortcut strategies, and is affected by issues such as dataset leakage, unanswerability, and benchmark saturation. They further demonstrate that model performance is sensitive to evaluation design choices and advocate for incorporating generative answering formats, improved scoring schemes, and principled test construction methods (e.g., Item Response Theory) to better assess model knowledge and robustness.  \citet{discriminative_muti_choice} proves that models fine-tuned for MCQA given only the set of answers without the question, are significantly outperforming the random accuracy baseline.

Another limitation of MCQA-based evaluation is the presence of systematic biases in LLM predictions. Selection bias refers to a model’s tendency to favor specific answer identifiers (e.g., A/B/C/D) regardless of their semantic content, even when the underlying text is modified \cite{zheng2024large}. Positional bias captures sensitivity to the ordering and placement of options within a question, independent of their meaning \cite{pezeshkpour2023large, zheng2024large}. More broadly, label bias arises when models prefer certain labels due to prompt-specific factors, such as label verbalization or the structure of in-context examples, rather than task-relevant information \cite{reif2024beyond}.

\subsection{Medical MCQA evaluation}

A substantial portion of research on medical benchmark construction has focused on MCQA datasets, including MedQA \cite{jin2021disease}, MedMCQA \cite{pal2022medmcqa}, and PubMedQA \cite{jin2019pubmedqa}, while comparatively fewer works examine the limitations of MCQA specifically from an evaluation perspective. Although these benchmarks have been instrumental in measuring progress, they often fail to reflect real clinical complexity, lack rigorous expert validation, and are increasingly saturated. Consequently, models tend to rely on pattern recognition rather than robust reasoning, sometimes correctly predicting answers even when key parts of the question are missing \cite{gu2025illusion, singh2025pitfalls}.

To address these issues, alternative evaluation paradigms based on free-text generation or multi-turn interactions have been proposed as more realistic and informative \cite{arora2025healthbench, singh2025pitfalls}. However, such approaches remain difficult to evaluate at scale due to their reliance on expert annotation, and while LLM-as-a-Judge methods offer partial automation, expert oversight is still required in safety-critical domains like healthcare \cite{szymanski2025limitations, diekmann2025llms}.

\subsection{Polish MCQA benchmarks}

LLMs are increasingly being explored in Polish-language medical settings, ranging from medical benchmarks to systems designed to support emergency care \cite{chojnicki2025pilot}. Selected specializations, question sets, and exam subsets from the Medical Final Exam (LEK, \textit{Lekarski Egzamin Końcowy}), the Medical–Dental Final Exam (LDEK, \textit{Lekarsko-Dentystyczny Egzamin Końcowy}), and the Polish Board Certification Examination (PES, \textit{Państwowy Egzamin Specjalizacyjny}) have been used in multiple studies to evaluate the LLMs' medical capacity in the Polish language \cite{rosol2023evaluation, wojcik2024comparative, suwala2023chatgpt, nicikowski2024potential, pokrywka2024gpt, siebielec2024assessment, jassem2025llmzsz}. 

The most comprehensive analysis to date of multiple LLMs on official Polish medical examinations is provided by \citet{grzybowski2025polish}. This study aggregates and publicly releases the largest collection of available exam materials. However, existing evaluations have not yet incorporated the Pharmaceutical Specialist Examination (PESF, \textit{Państwowy Egzamin Specjalizacyjny Farmaceutów}) or the Examination for Laboratory Diagnosticians’ Specialization (PESDL, \textit{Państwowy Egzamin Specjalizacyjny Diagnostów Laboratoryjnych}). To our knowledge, there are no studies evaluating Polish LLMs’ professional medical knowledge using open-ended responses.

\section{Polish Medical Exams} \label{sec:exams}

LEK and LDEK are Polish final examinations for medical and dental graduates. PES is a board certification examination for physicians and dentists after specialization training. PESDL and PESF are analogous specialist examinations for laboratory diagnosticians and pharmacists. All of them are multiple-choice exams with A–E answer options. They are described in more detail in Appendix~\ref{exams}.

Most exam questions are presented in the form of clinical vignettes reflecting realistic diagnostic and therapeutic scenarios. Such tasks often require interpretation of symptoms, laboratory findings, imaging results, and treatment options, thereby assessing applied clinical reasoning in addition to factual medical knowledge (example questions in Appendix~\ref{sec:modification-examples}).

\section{Dataset creation} \label{sec:dataset_creation}

In this section we describe how we collected additional exam data and modified the question format.

\subsection{Additional datasets sourcing}

We included PESF and PESDL exams using scraping scripts that extract questions and answer keys from PDFs. Two PESF editions were discarded due to OCR errors from scanned PDFs. We also added the most recent available exam editions. In total, this adds over 15,000 questions to the benchmark.

\subsection{Question modification} 
The MCQA format has inherent limitations when used to evaluate the actual knowledge of LLMs. Given a question \emph{Q} and a set of answer choices \emph{A}, including the correct answer $a_i \in A$, the task is often reduced to distinguishing the correct option from incorrect ones, rather than generating the correct answer independently.

To address this, we define three groups of question structures that can be modified to better reflect clinical reasoning. The examples of modified questions in the English version are provided in Appendix~\ref{sec:modification-examples}. These changes make questions more clinically oriented by reducing reliance on fixed answer choices.

\paragraph{1) Hidden multiple-response questions}
In the Multiple Statements (\textbf{MS}) setting, questions originally formulated as multiple-choice with composite answer options (each option encoding a subset of statements) are reformulated to require the model to output the set of correct statements directly. An answer is considered correct only if the predicted set exactly matches the ground truth.

In the Multiple Answers (\textbf{MA}) setting, questions containing meta-options (e.g., “answers C and D are correct”) are transformed into standard multiple-response questions by removing such options and marking the underlying answers as correct. Models must identify all of the answers to be correct.

\paragraph{2) Questions where the correct answer can be substituted with "None of the answers is correct."}
Correct answer substitution (\textbf{AS}) tests a model’s ability to abstain when no provided option is correct. In this setting, the original correct answer is replaced with an option indicating that none of the remaining answers is correct, requiring the model to recognize the absence of a valid choice.

\paragraph{3) Questions that can be modified to be open-ended.}
Open-ended questions (\textbf{OE}) are derived from exam items that can be answered solely based on their textual context, without reliance on predefined options. This format reduces multiple-choice bias and better reflects realistic model usage. However, evaluation is more challenging, as multiple valid phrasings may exist and domain expertise is required for judgment. In this setting, we employ LLM-as-a-judge.

The structure of the exam questions is consistent, enabling the filtering of questions into modification groups using regular expressions. The regular expression analysis and filtering were done in Polish. Since the English question sets are parallel to Polish, the same filters have been applied to them, as the question structures were the same.

\begin{table}[ht]
\centering
\resizebox{\linewidth}{!}{%
\begin{tabular}{lrrrr}
\toprule
Exam & \#Exam editions (\textit{Years}) & \#Original& \#Covered& Coverage \\
\midrule
LEK       & 26 (\textit{'08--'15, '20--'26}) & 4,728  & 2,875  & 61\% \\
LDEK      & 26 (\textit{'08-'15, '20--'26})            & 4,773  & 3,401  & 71\% \\
PES       & 12  (\textit{'08,'12,'15--'20,'23,'24})           & 9,965  & 7,070  & 71\% \\
PESDL     &   25 (\textit{'08--'20})             & 10,908 & 7,627  & 70\% \\
PESF      &   22 (\textit{'08--'20})             & 1,710  &   982  & 57\% \\
LEK EN   &    18 (\textit{'13--'15, '20--'26})            & 3,151  & 1,899  & 60\% \\
LDEK EN  &  18 (\textit{'13--'15, '20--'26})              & 3,196  & 2,223  & 70\% \\
\midrule
All       &                & 38,431 & 26,077 & 68\% \\
\bottomrule
\end{tabular}
}
\caption{Coverage statistics by source.}
\label{tab:coverage_stats}
\end{table}

\begin{figure*}[t]
    \centering
    \includegraphics[width=0.9\linewidth]{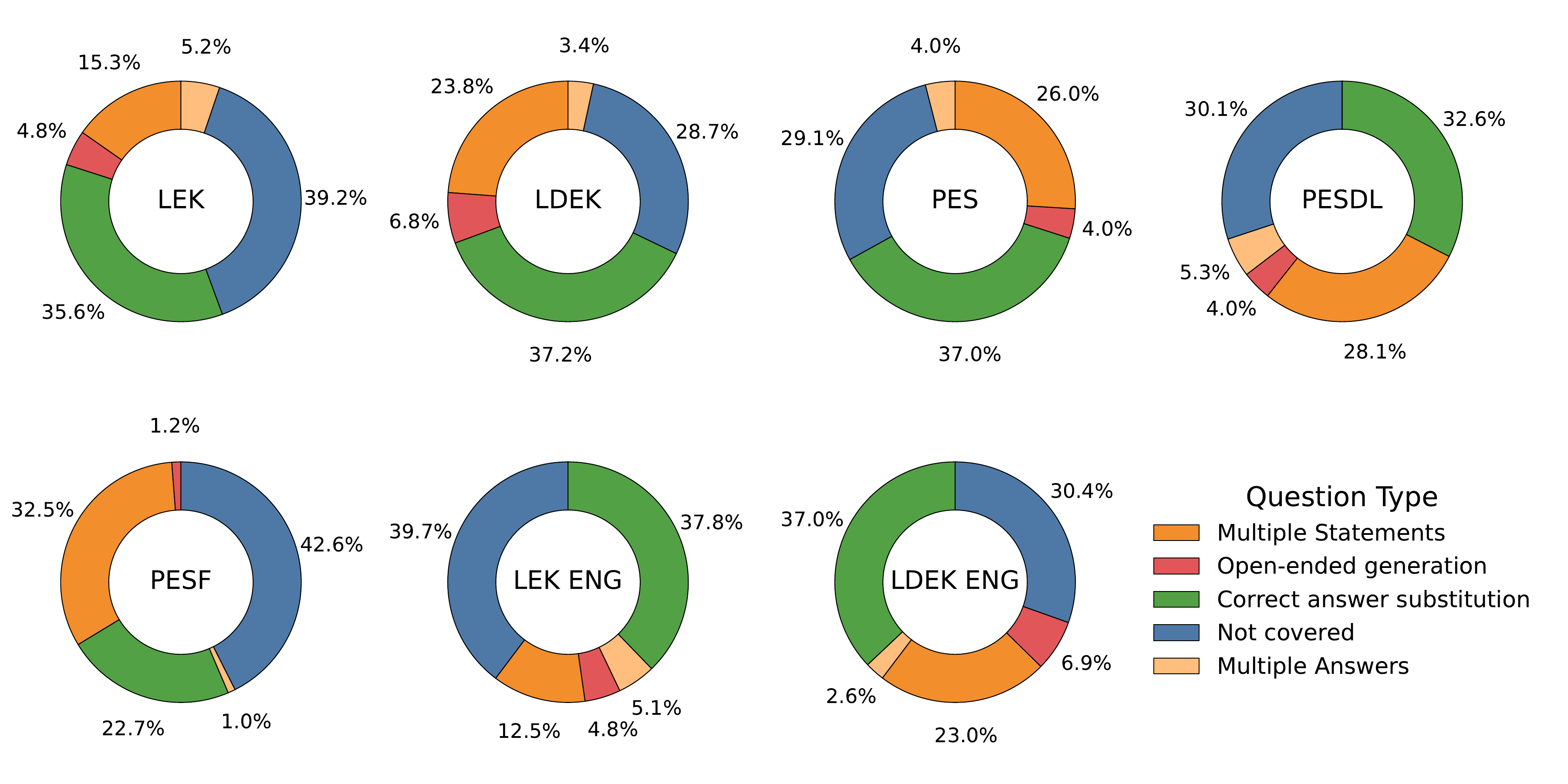}
    \caption{Question type composition across exam question sets.}
    \label{fig:composition}
\end{figure*}

The obtained dataset spans over 38,000 questions covers multiple medical domains (see Table~\ref{tab:coverage_stats} for details), and proposed alterations cover 68\% of the original questions. The distributions of altered question types vary across exam sets (as shown in Figure \ref{fig:composition}), because the questions suited for modifications are not distributed equally in each exam set. Additionally, open-ended alteration had priority over others, since it had the most rigorous filtering, so if the question was suited for both open-ended and \emph{None of the answers is correct} alterations, the open-ended was chosen.  

\section{Methodology} \label{sec:methodology}
\subsection{Obtaining answers}

Since LLMs are extremely vulnerable to prompt formulations, we have decided to inspect whether the prompt formulation strategy taken in \cite{grzybowski2025polish}, does not influence the distribution of the produced answers on the Polish medical exams benchmark. To achieve this, analysis of the produced answers by various LLMs, including Polish Bielik \cite{ociepa2025bielik11bv3, ociepa2025bielikv3smalltechnical} and PLLuM \cite{kocon2025pllum} variants, but also English-centered models like Qwen \cite{qwen3technicalreport} and GPT-5-mini \cite{openai2025gpt5mini} has been conducted using the instruction with example output format, which contained a sample answer (later denoted as \textbf{Previous Metodology (PM)}.

The initial analysis suggests that the considered models are biased towards the answer that is included as a sample answer format in the prompt, and that even the order of listing the answers in the instruction might influence the model's prediction. These initial takeaways are based on the comparison between the ground-truth answer distribution (correct answers for the medical exams) and the distribution of answers generated by the aforementioned models. Since the formulation of the original questions that comprise the benchmark already includes the possible answers in a classical order (A, B, C, D, E), providing the possible answers in the instruction is not necessary. Simple instruction paired with constrained generation limits this answer bias. This approach is later denoted as \textbf{Our evaluation Methodology (OM)}.

Our analysis shows that models are biased toward the example answer format, and that even answer ordering in the prompt can influence predictions. Comparing model outputs with the ground-truth distribution suggests systematic distortion introduced by the prompting scheme. Since the benchmark already provides options in a fixed A--E format, explicitly repeating them in the prompt is unnecessary. We therefore adopt a simpler instruction setup with constrained generation to reduce prompt-induced bias, referred to as \textbf{Our Evaluation Methodology (OM)}.

When applicable\footnote{Some of the tokenizers have unique way of encoding the numbers into multiple tokens, when constrained to generate only tokens that are used to form numbers from 1-20 (indices of correct statements) the model generated complex multi-digit numbers instead, which rendered a constrained generation unapplicable.}, constrained generation based on a set of allowed tokens was used paired with the instruction suited for the modified and original question types. 
For example, in \textbf{MS} scenario, the model had to return the set of correct statements; the generation was constrained to return only the indices of the statements considered true, and a suitable instruction was modified to meet these requirements. The refined instruction and all of the instructions for generating answers for modified questions are provided in the Appendix \ref{sec:instructions}.

\subsection{Answer evaluation}
\paragraph{Answer-key}
During the dataset alteration, the answer key for a given modification was modified on the fly to resemble the modifications made. When considering multi-answer or multi-statement questions, the answer was deemed only correct when the provided output included an identical set of correct statement indices or answers. 

\paragraph{LLM-as-a-judge}
For evaluating open-ended questions, the LLM-as-a-judge approach was adopted using Deepseek-V4-Pro \cite{deepseekai2026deepseekv4}. However, since the original answer to the modified question may not be the only correct one, and judgment based solely on the original answer key would be insufficient, the model requires extensive medical knowledge to make a correct assessment. Additionally, binary judgments can overstate the accuracy of LLMs as evaluators, since the two-class setup allows even chance-level predictions to score relatively high. Instruction for the LLM judgement is provided in the Appendix \ref{sec:instructions}.

\paragraph{Annotations}
We build a human meta-evaluation set to test whether LLM-as-a-judge aligns with medical experts and whether out-of-domain LLMs are reliable for evaluation. We sample LEK and PES open-ended questions and collect answers from four models: PLLuM-12B-instruct \cite{kocon2025pllum}, gpt-oss-20b \cite{openai2025gptoss120bgptoss20bmodel}, Bielik-11B-v2.6-Instruct \cite{ociepa2025bielik11bv2technical}, and Qwen3-30B-A3B-Instruct-2507 \cite{qwen3technicalreport}. Six medical experts annotated 200 (question, answer, reference answer) triplets each, with 15 control items per set. The task was binary: decide whether the generated answer is correct, with the reference answer as context; annotators were also able to discard unclear or invalid questions.

In total, we obtained 927 valid annotations and 90 control samples (see Table \ref{tab:annotation_stats}); over 150 items were removed due to poor or outdated questions. We computed inter-annotator agreement (Fleiss’ $\kappa$) on controls and removed one outlier annotator. We then measured agreement between experts and the LLM judge using Cohen’s $\kappa$, obtaining 0.649, which indicates substantial agreement but shows that LLM-based evaluation is not equivalent to expert judgment. Annotation and LLM judge guidelines are aligned across both setups.

\begin{table}
\centering
\resizebox{\linewidth}{!}{%
\begin{tabular}{lcc}
\toprule
& \rotatebox{45}{Human-Human} 
& \rotatebox{45}{Human-Model} \\
\midrule
All Annotation Samples & 1,200 & --- \\
Invalid Samples -- Missing Labels & 16 & --- \\
Invalid Samples -- Invalid Question & 167 & --- \\
Invalid Samples -- Annotator - Outlier & 184 & --- \\
\midrule
Valid samples & \multicolumn{2}{c}{--- 743 ---} \\
Positive Labels & 220 & 270 \\
Negative Labels & 523  & 473\\ 
\midrule
Kappa & \shortstack{0.921 \\(\textit{Fleiss - w/o outlier})} & \shortstack{0.649 \\ (\textit{Cohen})} \\
\bottomrule
\end{tabular}
}
\caption{Annotation process summary}
\label{tab:annotation_stats}
\end{table}

\paragraph{Disagreement analysis}

We analysed disagreements between human annotators and LLM-as-a-judge models to identify common sources of divergence. A senior physician reviewed a subset of 74 cases with inconsistent judgments and identified six recurring \textbf{discrepancy patterns} (Table~\ref{tab:discrepancy_patterns_overview}).

Further annotation of these cases showed that disagreements mainly arise from two factors: (1) linguistic issues (e.g., poor Polish fluency, incorrect terminology), and (2) differences in judging correctness. Human annotators tend to focus on clinically or linguistically relevant errors, while LLM judges rely more on surface overlap with the reference answer and often miss contextual or off-target errors.
Additional details are provided in Appendix~\ref{sec:disagreement_analysis}.

\begin{table}
    \centering
    \resizebox{\linewidth}{!}{%
    \begin{tabular}{cp{2.5cm}p{7cm}}
        \toprule
        \multicolumn{2}{c}{\textbf{Discrepancy pattern}}& \textbf{Definition} \\
        \midrule
        A & Errors in redundant content & The answer included unnecessary additional content which introduced at least one defect.\\
        B & Valid alternative interpretation of the question & The answer addressed a valid interpretation of the open-ended question, but not the issue implied by the original closed-ended question and answer key.\\
        C & Underspecified expected detail & The answer addressed the expected issue, but with less specificity than the official correct answer.\\
        D & Non-existent medical terms & The answer used non-existent medical terms, including hybrid Polish–English or Polish–Latin expressions.\\
        E & Misuse of existing medical terms & The answer used existing medical terms  inconsistently with their accepted meaning.\\
        F & Incorrect spelling or notation of medical terms & The answer used existing medical terms with incorrect spelling or notation.\\
        \bottomrule
    \end{tabular}
    }
    \caption{Overview of discrepancy patterns used to label potential sources of disagreement between human annotators and AI judges}
    \label{tab:discrepancy_patterns_overview}
\end{table}

\section{Experiment setup} \label{sec:experiment_setup}
The experiments consist of two parts. The first part focuses on the outcomes of the efforts to limit the bias and create a fair evaluation. In this part, the LLMs are evaluated on the unmodified exam sets, comparing two evaluation techniques: the technique reproduced from the Polish Medical Exams benchmark \cite{grzybowski2025polish}  and the introduced one that changes the instruction to minimize bias and utilizes constrained generation. Additionally, including the PESF and PESDL exam questions as an extension to the benchmark. 
The second part of the experiments presents a comparison of the results obtained from the evaluation using the original and altered question structures. Each exam set is divided according to the alteration method and then evaluated on the same questions, both unmodified and altered.

\subsection{Models}
Models under study are categorized in the following way: medical-domain fine-tuned models, Polish LLMs, and general-purpose LLMs.

\paragraph{Medical models:} BioMistral-7B and BioMistral-7B-DARE \cite{labrak2024biomistral} , Meditron3-70B \cite{OpenMeditron_Meditron3_70B}, Llama3-OpenBioLLM-70B \cite{OpenBioLLM_70B}, MedGemma (4B and 27B versions) \cite{sellergren2025medgemmatechnicalreport}, MediPhi \cite{mediphi}. 

\paragraph{General-purpose models:} 
Qwen3.5 in three sizes 9B, 35B and 122B \cite{qwen3.5}, Llama3.3-70B \cite{meta_llama_3_3_70b_instruct}, Llama3-8B \cite{llama3modelcard}, Gemma3 (12B and 27B versions) \cite{gemma_2025}.

\begin{table}[t]
\centering
\small
\setlength{\tabcolsep}{7pt}
\renewcommand{\arraystretch}{1.08}
\resizebox{1.0\linewidth}{!}{%
\begin{tabular}{l S[table-format=2.1] S[table-format=2.1] r}
\toprule
\textbf{Model} & {\textbf{Avg OM}} & {\textbf{Avg PM}} & {$\boldsymbol{\Delta}$ \tiny{\textbf{(OM$-$PM)}}}
\\
\midrule
\multicolumn{4}{l}{\hphantom{-x-x-}\textit{Polish LLMs}} \\
\midrule
Bielik-4.5B          & 37.8 & 36.8 & \textcolor{ForestGreen}{+1.1}  \\
Bielik-11B           & 55.8 & \textbf{55.0} & \textcolor{ForestGreen}{+0.8}  \\
PLLuM-12B-NC-Inst    & 37.3 & 36.5 & \textcolor{ForestGreen}{+0.8}  \\
PLLuM-12B-NC-Chat    & 38.1 & 37.3 & \textcolor{ForestGreen}{+0.7}  \\
PLLuM-12B-Inst       & 35.9 & 27.5 & \textcolor{ForestGreen}{+8.4}  \\
PLLuM-12B-Chat       & 29.5 & 17.9 & \textcolor{ForestGreen}{+11.6} \\
PLLuM-70B-Inst       & \textbf{57.1} & 39.4 & \textcolor{ForestGreen}{+17.7} \\
\midrule
\multicolumn{4}{l}{\hphantom{-x-x-}\textit{General LLMs}} \\
\midrule
LLaMA3-8B            & 41.8 & 43.3 & \textcolor{BrickRed}{-1.4}    \\
Qwen3.5-9B           & 63.1 & 42.9 & \textcolor{ForestGreen}{+20.2} \\
Gemma-3-12B          & 53.3 & 44.5 & \textcolor{ForestGreen}{+8.9}  \\
Gemma-3-27B          & 59.2 & 55.9 & \textcolor{ForestGreen}{+3.3}  \\
Qwen3.5-35B-A3B        & 72.2 & 71.0 & \textcolor{ForestGreen}{+1.2}  \\
LLaMA3.3-70B         & 65.8 & 66.4 & \textcolor{BrickRed}{-0.6}   \\
Qwen3.5-122B-A10B    & \underline{\textbf{75.1}} & \underline{\textbf{75.8}} & \textcolor{BrickRed}{-0.7}  \\
\midrule
\multicolumn{4}{l}{\hphantom{-x-x-}\textit{Medical LLMs}} \\
\midrule
MedGemma-4B          & 43.4 & 38.0 & \textcolor{ForestGreen}{+5.3}  \\
MediPhi-Instruct     & 37.0 & 31.2 & \textcolor{ForestGreen}{+5.8}  \\
BioMistral-7B        & 30.6 & 21.1 & \textcolor{ForestGreen}{+9.6}  \\
BioMistral-7B-DARE   & 30.7 & 24.3 & \textcolor{ForestGreen}{+6.4}  \\
MedGemma-27B        & \textbf{60.0} & \textbf{56.9} & \textcolor{ForestGreen}{+3.1}  \\
OpenBioLLM-70B       & 57.2 & 50.2 & \textcolor{ForestGreen}{+7.0}  \\
Meditron3-70B        & 55.8 & 16.8 & \textcolor{ForestGreen}{+38.9} \\
\bottomrule
\end{tabular}
}%
\caption{%
  Average model performance under the proposed bias-mitigated methodology (OM) and
  the previous MCQA protocol (PM).
  Averages computed over all available exams (LEK, LDEK, PES, PESDL, PESF,
  LEK-EN, LDEK-EN).
  Best result per model family in \textbf{bold}, best overall is \underline{underlined}. Full per-exam breakdown in Table~\ref{tab:short_bias_nobias}.
}
\label{tab:avg_summary_bias}
\end{table}

\begin{table}[t]
\centering
\small
\setlength{\tabcolsep}{7pt}
\renewcommand{\arraystretch}{1.08}
\resizebox{1.0\linewidth}{!}{%
\begin{tabular}{l S[table-format=2.2] r r r}
\toprule
\textbf{Model} & {\textbf{Avg Mod}} & \textbf{Avg Og}& {\textbf{$\boldsymbol{\Delta}$ EN}} & {\textbf{$\boldsymbol{\Delta}$ PL}} \\
\midrule
\multicolumn{5}{l}{\hphantom{-x-x-}\textit{Polish LLMs}} \\
\midrule
Bielik-4.5B                  & 13.01 & 36.30&  \textcolor{BrickRed}{-27.64} & \textcolor{BrickRed}{-21.56} \\
Bielik-11B                   & \textbf{17.49}& 56.27 & \textcolor{BrickRed}{-40.64} & \textcolor{BrickRed}{-38.04} \\
PLLuM-12B-NC-Inst            & 10.65 & 36.39& \textcolor{BrickRed}{-27.94} & \textcolor{BrickRed}{-24.87} \\
PLLuM-12B-NC-Chat            & 9.50 & 37.22 & \textcolor{BrickRed}{-30.40} & \textcolor{BrickRed}{-26.66} \\
PLLuM-12B-Inst               & 8.03 & 34.71 &\textcolor{BrickRed}{-27.92} & \textcolor{BrickRed}{-26.18} \\
PLLuM-12B-Chat               & 6.72 & 29.11 &\textcolor{BrickRed}{-20.24} & \textcolor{BrickRed}{-23.26} \\
PLLuM-70B-Inst               & 14.62 & \textbf{57.04} &\textcolor{BrickRed}{-45.26} & \textcolor{BrickRed}{-41.29} \\
\midrule
Average & 11.43 & 41.01 &\textcolor{BrickRed}{-31.43} & \textcolor{BrickRed}{-28.84} \\
\midrule \midrule
\multicolumn{5}{l}{\hphantom{-x-x-}\textit{General LLMs}} \\
\midrule 
LLaMA3-8B                    & 19.12 &  41.27&\textcolor{BrickRed}{-26.92} & \textcolor{BrickRed}{-20.25} \\
Qwen3.5-9B                   & 21.93 & 64.43&  \textcolor{BrickRed}{-42.38} & \textcolor{BrickRed}{-42.55} \\
Gemma-3-12B                  & 9.08 & 54.35 & \textcolor{BrickRed}{-41.04} & \textcolor{BrickRed}{-46.96} \\
Gemma-3-27B                  & 19.92 & 59.96& \textcolor{BrickRed}{-41.56} & \textcolor{BrickRed}{-39.43} \\
Qwen3.5-35B                  & 39.98 &  73.96 &\textcolor{BrickRed}{-37.59} & \textcolor{BrickRed}{-32.53} \\
LLaMA3.3-70B                 & 22.92 & 67.14 & \textcolor{BrickRed}{-47.82} & \textcolor{BrickRed}{-42.78} \\
Qwen3.5-122B                 & \underline{\textbf{46.97}} & \underline{\textbf{77.28}} & \textcolor{BrickRed}{-28.49} & \textcolor{BrickRed}{-31.04} \\
\midrule
Average & 25.70 & 62.63 & \textcolor{BrickRed}{-37.97} & \textcolor{BrickRed}{-36.51} \\
\midrule \midrule
\multicolumn{5}{l}{\hphantom{-x-x-}\textit{Medical LLMs}} \\
\midrule
MedGemma-4B                  & 9.10 & 42.49 & \textcolor{BrickRed}{-36.08} & \textcolor{BrickRed}{-32.33} \\
BioMistral-7B                & 5.49 & 30.02 & \textcolor{BrickRed}{-29.36} & \textcolor{BrickRed}{-22.60} \\
BioMistral-7B-DARE           & 7.54 & 29.48 & \textcolor{BrickRed}{-30.25} & \textcolor{BrickRed}{-18.62} \\
MediPhi-Instruct             & \textbf{31.87} & 37.77 & \textcolor{BrickRed}{-30.16} & \textcolor{ForestGreen}{+3.80} \\
MedGemma-27B                 & 14.14 &  \textbf{60.78} &\textcolor{BrickRed}{-52.01} & \textcolor{BrickRed}{-44.49} \\
OpenBioLLM-70B               & 26.66 & 57.41 &\textcolor{BrickRed}{-32.22} & \textcolor{BrickRed}{-30.17} \\
Meditron3-70B                & 25.93 & 56.41 &\textcolor{BrickRed}{-42.51} & \textcolor{BrickRed}{-25.45} \\
\midrule
Average & 17.25 & 44.89& \textcolor{BrickRed}{-36.08} & \textcolor{BrickRed}{-24.27} \\
\bottomrule
\end{tabular}
}%
\caption{%
  Average model performance on modified questions (Avg Mod), original variants of the same questions (Avg Og) and deltas between the performance on modfied questions and original questions on English ($\Delta$~EN) exams (LEK-EN, LDEK-EN)
  and Polish ($\Delta$~PL) exams (LEK, LDEK, PES, PESDL, PESF).
  \textbf{Bold} values indicate highest scores within model group, while highest value overall is \underline{underlined}. Full per-exam breakdown in Table~\ref{tab:avg_delat}.
}
\label{tab:avg_delat_short}
\end{table}

\begin{table}
\centering
\resizebox{0.5\textwidth}{!}{
\begin{tabular}{lcccccccc}
\toprule
 & \multicolumn{8}{c}{\textbf{Open-ended question accuracies (\%)}} \\
\cmidrule(lr){2-9}
\textbf{Model} 
& \rotatebox{60}{LEK} 
& \rotatebox{60}{LDEK} 
& \rotatebox{60}{PES} 
& \rotatebox{60}{PESDL} 
& \rotatebox{60}{PESF} 
& \rotatebox{60}{LEK-EN} 
& \rotatebox{60}{LDEK-EN} 
& \rotatebox{60}{Avg} \\
\midrule

\multicolumn{9}{l}{\hphantom{-x-x-}\textit{Polish LLMs}} \\
\midrule
Bielik-4.5B        & 34.67 & 21.23 & 14.14 & 22.92 & 4.76  & 29.61 & 19.46 & 20.97 \\
Bielik-11B         & \textbf{58.22} & \textbf{38.77} & \textbf{31.06} & 38.43 & 38.10 & 59.87 & 33.94 & 42.63 \\
PLLuM-12B-NC-Inst  & 48.44 & 31.38 & 22.73 & 36.34 & 28.57 & 36.84 & 28.05 & 33.19 \\
PLLuM-12B-NC-Chat  & 54.22 & 32.31 & 25.25 & 35.19 & 14.29 & 42.76 & 25.34 & 32.77 \\
PLLuM-12B-Inst     & 41.33 & 24.62 & 19.95 & 26.85 & 19.05 & 41.45 & 24.43 & 28.24 \\
PLLuM-12B-Chat     & 43.11 & 24.00 & 16.41 & 24.77 & 19.05 & 37.50 & 20.81 & 26.52 \\
PLLuM-70B-Inst     & 54.67 & 36.62 & 26.52 & \textbf{45.37} & \textbf{52.38} & \textbf{63.16} & \textbf{46.15} & \textbf{46.41} \\

\midrule
\multicolumn{9}{l}{\hphantom{-x-x-}\textit{General LLMs}} \\
\midrule
LLaMA3-8B          & 25.78 & 18.77 & 15.66 & 22.69 & 9.52  & 39.47 & 32.58 & 23.50 \\
Qwen3.5-9B         & 45.78 & 32.92 & 25.25 & 39.81 & 42.86 & 63.16 & 48.87 & 42.66 \\
Gemma-3-12B        & 45.33 & 29.54 & 26.77 & 29.63 & 23.81 & 54.61 & 33.48 & 34.74 \\
Gemma-3-27B        & 56.00 & 40.92 & 33.59 & 46.76 & 52.38 & 54.61 & 43.44 & 46.81 \\
Qwen3.5-35B-A3B    & 63.56 & 42.77 & 38.64 & 55.79 & 47.62 & 65.79 & 52.04 & 52.32 \\
LLaMA3.3-70B       & 60.89 & 36.00 & 30.05 & 46.76 & 33.33 & 62.50 & 48.87 & 45.49 \\
Qwen3.5-122B-A10B  & \underline{\textbf{68.00}} & \underline{\textbf{53.54}} & \underline{\textbf{52.78}} & \underline{\textbf{61.11}} & \underline{\textbf{76.19}} & \underline{\textbf{73.03}} & \underline{\textbf{57.01}} & \underline{\textbf{63.09}} \\

\midrule
\multicolumn{9}{l}{\hphantom{-x-x-}\textit{Medical LLMs}} \\
\midrule
MedGemma-4B        & 27.11 & 18.46 & 12.63 & 17.36 & 23.81 & 41.45 & 22.62 & 23.35 \\
MediPhi-Instruct   & 6.67  & 5.85  & 4.80  & 7.41  & 9.52  & 40.13 & 33.48 & 15.41 \\
BioMistral-7B      & 16.00 & 8.92  & 6.06  & 9.95  & 4.76  & 36.84 & 26.70 & 15.60 \\
BioMistral-7B-DARE & 22.67 & 13.85 & 7.32  & 13.66 & 9.52  & 36.18 & 27.15 & 18.62 \\
MedGemma-27B       & \textbf{56.44} & 33.54 & \textbf{33.33} & \textbf{44.68} & 33.33 & 56.58 & 45.25 & 43.31 \\
OpenBioLLM-70B     & 55.11 & \textbf{35.08} & 28.79 & 41.90 & 23.81 & 57.24 & \textbf{46.61} & 41.22 \\
Meditron3-70B      & 51.11 & 34.46 & 32.07 & 42.36 & \textbf{52.38} & \textbf{66.45} & \textbf{46.61} & \textbf{46.49} \\

\bottomrule
\end{tabular}
}
\caption{Performance of models on open-ended questions across exams as the percentage of correctly generated answers judged by the Deepseek-V4-Pro. ``\textit{Avg}'' column is the mean across accuracies on all exams. \textbf{Bold} indicates the best performer within each model group, while \underline{underline} denotes the best score overall.}
\label{tab:open_ended_generation}
\end{table}

\paragraph{Polish LLMs:} 
PLLuM-12B-instruct, Llama-PLLuM-70B-instruct \cite{kocon2025pllum}, Bielik-11B-v3-Instruct \cite{ociepa2025bielik11bv3} and Bielik-4.5B-v3.0-Instruct \cite{ociepa2025bielikv3smalltechnical}

\subsection{Evaluation}
We use accuracy expressed in \% as a metric for evaluation. Additionally, we aggregate the accuracies from respective modification groups, taking an average of \textbf{MS}, \textbf{MA} and \textbf{AS} and an accuracy on the unmodified equivalents to measure the average change of the performance of a given model on a given exam between the original and modified questions. The evaluation of the \textbf{OE} questions is also expressed as an accuracy measure where the correct answer is determined based on the judgment from the described LLM-as-a-judge approach using Deepseek-V4-Pro \cite{deepseekai2026deepseekv4}.

\section{Results and discussion} \label{sec:results_and_discussion}
\paragraph{Limiting the bias}
Table~\ref{tab:avg_summary_bias} compares the replicated evaluation protocol from \citet{grzybowski2025polish} with the proposed bias-mitigated methodology. Most models achieve better results under the proposed protocol, which supports the hypothesis that the original instruction and generation setup introduces bias that lowers benchmark scores. The difference is especially visible for some instruction-tuned models, such as PLLuM-70B-Inst, Qwen3.5-9B, and Meditron3-70B.

The proposed protocol also changes the ranking of the Polish LLMs. Under the bias-mitigated evaluation, PLLuM-70B-Inst achieves a higher average score than Bielik-11B, while the previous methodology suggests the opposite ranking. 
This suggests that the original protocol may affect some models more strongly than others. 
Overall, the results indicate that limiting instruction-related bias provides a more reliable comparison between models.

\paragraph{Question alteration effect}
Table \ref{tab:avg_delat_short} presents the differences in accuracy between the original and altered questions. The significant performance drop observed across models and examinations in both Polish and English indicates that model performance is highly dependent on the evaluation format. General-purpose models achieved the strongest results under the altered conditions, with Qwen3.5-122B obtaining the best overall performance.

The alterations expose the vulnerabilities of models, showing that under the changed conditions they fail to provide the correct answer, especially in lower-resource language like Polish, where the average performance on modified questions is particularly low. This only shows how it is important to create a robust and reliable evaluation for lower-resource languages in the critical domains like medicine.  

The only model that experienced gains in performance over the refined evaluation was MediPhi-Instruct, becoming the best performer in the Medical LLMs group. However, under further investigation the source of the improvement was found in the questions from the \textbf{AS} question group on Polish exams (see Appendix \ref{sec:detailed_results} Table \ref{tab:corr_ans_sub_vs_original} and Appendix \ref{app:correct_ans_sub}). The MediPhi inconsistency between performance on English and Polish within the same alteration group can be attributed to the model's overfitting towards the "None of the answers is correct" option.

\paragraph{Open-ended questions}
Answering open-ended questions is the most natural setting for evaluating LLMs, as it resembles both day-to-day and in-domain use. The results presented in Table \ref{tab:open_ended_generation} show that, on average, the evaluated LLMs fail to correctly answer expert-level exam questions. Medical LLMs, which are expected to perform particularly well in this setting, achieved results comparable to those of general-purpose Polish LLMs trained holistically rather than specifically for medical tasks. 

The relatively low performance on the question pool spanning multiple medical specialisations and domains, since this setting is closest to the day-to-day use of the LLMs by regular consumers. 

\paragraph{Model groups}
Among models with comparable parameter counts across the groups, we observe that the Polish PLLuM-70B experienced substanstial performance drop on the altered questions compared to general LLaMA-3.3-70B, or medical Meditron3 or OpenBioLLM, which performed the best for their size. On open-ended questions, these models perform comparably, but Meditron3 outperforms the others by a slight margin. 

These comparisons are complex since the models need to be fluent in English, Polish, and have expert-level medical knowledge. Additionally, the reformulated evaluation is more demanding since it does not follow the classic MCQA approach.

\paragraph{Data contamination}
\begin{table}[th]
\centering
\label{tab:contamination}
\resizebox{\columnwidth}{!}{
\begin{tabular}{lrrr}
\hline
\textbf{Model} & \textbf{2015} & \textbf{2020} & \textbf{2026} \\
\hline
Bielik-11B & (17.0, 24.5)  & (13.5, 19.0)  & (12.5, 23.5) \\
PLLuM-12B-Inst      & (13.5, 19.0)  & (10.37, 26.5) & (18.0, 20.0) \\
\midrule
LLaMA-3.3-70B   & (17.5, 18.0)  & (18.5, 19.0)  & (13.5, 18.0) \\
Qwen3.5-35B           & (8.59, 9.5)   & (5.53, 6.0)   & (9.6, 10.5)  \\
\midrule
MedGemma-27B       & (6.63, 15.5)  & (3.3, 12.0)   & (4.74, 9.5)  \\
Meditron3-70B          & (9.5, 9.5)    & (10.5, 12.5)  & (11.0, 13.0) \\
\hline
\end{tabular}
}
\caption{Contamination level ranges reported using DCQ methodology across editions of LEK and LDEK, provided in format (min contamination, max contamination).}
\label{tab:cont}
\end{table}

Since the benchmark was constructed from publicly available data, exposure of the questions in LLM training corpora is possible. 
To investigate this, we conducted additional temporal contamination analysis using the DCQ method \cite{DCQ} on the Polish LEK and LDEK exams from 3 editions (2015, 2020, 2026), 200 questions each. 
The results suggest limited contamination (see Table \ref{tab:cont}), yet the reported levels do not align with the performance drops observed when applying question alterations. Notably, the DCQ measured exposure to exam questions only, which are considerably more accessible than their corresponding answers. Furthermore, if contamination were present, the levels reported for the 2026 edition would be expected to be substantially lower than those for the 2015 and 2020 editions -- a pattern that is not observed in the data.

Moreover, we have performed Web Search Audit which revealed limited exposure of LEK exam questions and no exposure of PES exam questions and analyzed the model's performance over the years on the English LEK and LDEK examinations (detailed results in Appendix~\ref{contamination}).

\section{Conclusion} \label{sec:conclusion}

We extended and refined a Polish medical exam benchmark to provide a more informative evaluation of LLMs beyond standard single-choice questions. 
Our results show that both question structure and domain coverage substantially influence model performance, suggesting that high scores may not reliably reflect true medical competence. These findings underscore the need for more robust, linguistically and contextually grounded evaluation frameworks as well as better pre-training protocols for medical LLMs.

\section*{Limitations}
While we present comprehensive evaluation of various models, due to computational and cost limitations, we did not exhaust all of the experimental possibilities, which could include more models in the same size range, another family of models like Mistral, or flagship commercial models like Claude or GPT5. 

The Deepseek-V4-Pro model used in the LLM-as-a-judge approach, although shows substantial agreement with expert annotations, is not specialised in the medical domain. 
This may limit its ability to consistently identify subtle clinical inaccuracies or domain-specific reasoning errors.

The regex approach employed to filter the questions into modification groups does not capture all edge cases, leaving the possibility of errors in generated answers due to mismatches between questions and instructions. 
Such errors are expected to be infrequent but could introduce noise into the analysis for certain modification types.
    
\section*{Ethical Considerations}

The exam questions used in this work were originally created by the Centre of Medical Examination (Centrum Egzaminów Medycznych - CEM) and made publicly available. Our contribution is limited to restructuring and modifying them. The part of our dataset was also developed based on an existing benchmark dataset \cite{grzybowski2025polish}, with permission from its authors, and we preserve attribution to the original sources.

To validate the quality of the LLM-as-a-judge setup, we used Polish-speaking human annotators with medical expertise. Annotators were recruited via direct personal invitations and collaborated outside any commercial crowd-annotation platform. This helped ensure domain competence and reduced risks related to low-quality or uninformed annotations.

Importantly, performance on written medical exams reflects only a narrow slice of medical competence. Clinical practice additionally requires, among other things, taking a patient history, performing a physical examination, interpreting laboratory and imaging results, considering contraindications and comorbidities, communicating uncertainty, and making decisions under incomplete information. Therefore, high benchmark scores should not be interpreted as evidence that LLMs “outperform doctors” or are ready to replace clinical professionals.

Finally, while LLMs can be useful for medical education and as decision-support tools, they may still produce hallucinated or incorrect outputs. This creates a safety risk if such outputs are treated as authoritative. We stress that clinicians should be explicitly informed about these limitations and encouraged to verify model outputs, particularly because medical errors can have serious consequences for patient health and life.

\section{Acknowledgments}
We gratefully acknowledge Polish high-performance computing infrastructure PLGrid (HPC Center: ACK Cyfronet AGH) for providing computer facilities and support within computational grant no. PLG/2026/019405.

\bibliography{custom}
\appendix

\section{Usage of GenAI in Research}

We used Grammarly and GPT models for improving the grammar of this manuscripts as well as ChatGPT for coding. All GenAI outputs were manually verified and accepted by the authors.

\section{Experimental Setup and Hyperparameters}
Experiments were conducted on a cluster consisting of NVIDIA A100 SXM4-40GB GPUs and lasted around 500 GPU/Hours. The maximum number of used GPUS for a single model evaluation was 4. To ensure reproducibility, the generation of the answers was done with a temperature equal to 0, constrained generation (when applicable) was limited to answers A, B, C, D, E, and integers from 1 to 20 with the stop token set to "." depending on the question form. 

For the data creation and formatting we have used standard packages: Pandas, Huggingface Datasets, re, pickle. For the LLMs evaluation, we were using vLLM and Huggingface Transformers with two separate environments: one for most of the models (vLLM 0.10.1.1, Transformers 4.56.0), the second one for the newest models from the Qwen3.5 series (vLLM 0.19.0, Transformers 5.5.4). For the LLM-as-a-judge testing and deployment: OpenAI, vLLM, and anthropic packages.  
For visualisations seaborne and matplotlib.

\section{Exams}
\label{exams}
The LEK Medical Final Exam in Poland is required for medical faculty graduates to obtain the right to practice and be admitted into a specialization. The exam consists of 200 closed questions with four distractors and one correct answer. Although a score of 56\% is sufficient to pass, placement into a specialization depends on ranking; therefore, most candidates aim for scores of around 90\%.  Similarly, LDEK is an equivalent of LEK but for dentistry graduates.
The PES Polish Board Certification Exam is a mandatory exam for physicians and dentists who have completed their specialization. It consists of two parts: a written exam and an oral. The written part consists of 120 closed questions with four distractors and one correct answer, and the participant passes if they obtain at least 60\% of the possible points. 
Laboratory diagnosticians and pharmacists have their own equivalents of the PES, the Pharmaceutical Specialist Examination (PESF), and the Examination for Laboratory Diagnosticians Specialization (PESDL). Both of them have the same structure and characteristics as PES.
Previous studies \cite{grzybowski2025polish} included from LEK, LDEK and PES together with English counterparts of LEK and LDEK, from the Centre of Medical Examination (Centrum Egzaminów Medycznych - CEM)\footnote{\url{https://cem.edu.pl/index.php}} and Supreme Medical Chamber (Naczelna Izba Lekarska - NIL).\footnote{\url{https://nil.org.pl/}} Recently, these institutions have released the question sets from the PESF and PESDL exams, which have not been included in any of the prior studies. This has opened an avenue for more diverse evaluation of medical knowledge.

\section{Detailed results}
Tables \ref{tab:original_vs_multistatement}, \ref{tab:tab:original_vs_multianswer} and \ref{tab:corr_ans_sub_vs_original} present the detailed results over the different question alteration groups. Each group of modified questions from a given exam is compared with the same set of questions, but unmodified, and an evaluation method with constrained generations is used. 

\label{sec:detailed_results}
\begin{table*}
\resizebox{\textwidth}{!}{
\begin{tabular}{lcccccccccccccc}
\toprule
 & \multicolumn{14}{c}{\textbf{Exam Performance (\%)}} \\
\cmidrule(lr){2-15}
\textbf{Model} &
\multicolumn{2}{c}{LEK} &
\multicolumn{2}{c}{LDEK} &
\multicolumn{2}{c}{PES} &
\multicolumn{2}{c}{PESDL} &
\multicolumn{2}{c}{PESF} &
\multicolumn{2}{c}{LEK-EN} &
\multicolumn{2}{c}{LDEK-EN} \\
\cmidrule(lr){2-3}\cmidrule(lr){4-5}\cmidrule(lr){6-7}
\cmidrule(lr){8-9}\cmidrule(lr){10-11}\cmidrule(lr){12-13}\cmidrule(lr){14-15}
 & OM & MS & OM & MS & OM & MS & OM & MS & OM & MS & OM & MS & OM & MS \\
\midrule
Bielik-4.5B           & 22.96 & 17.57 & 21.99 & 9.94 & 21.98 & 10.16 & 3.79 & 8.98 & 8.11 & 19.64 & 43.54 & 15.95 & 36.24 & 9.26 \\
Bielik-11B            & \textbf{62.79 }& \textbf{42.05} & \textbf{44.06} & \textbf{19.26} & \textbf{40.17} & 19.04 & 47.88 & 21.54 & \textbf{65.05} & \textbf{45.05} & \textbf{58.73} & 35.44 & \textbf{40.33} & 16.35 \\
PLLuM-12B-NC-Inst     & 33.75 & 22.41 & 23.66 & 10.73 & 25.26 & 11.9 & 26.89 & 12.6 & 29.55 & 31.71 & 22.28 & 9.62 & 18.12 & 5.59 \\
PLLuM-12B-NC-Chat     & 31.12 & 15.63 & 25.42 & 8.18 & 24.45 & 9.66 & 25.82 & 12.63 & 31.89 & 29.19 & 26.33 & 8.1 & 20.84 & 4.9 \\
PLLuM-12B-Inst        & 27.25 & 8.44 & 22.25 & 4.66 & 23.29 & 7.07 & 25.26 & 5.71 & 28.29 & 14.23 & 22.53 & 10.63 & 21.93 & 5.59 \\
PLLuM-12B-Chat        & 24.76 & 3.46 & 21.55 & 1.58 & 21.78 & 2.55 & 24.54 & 2.19 & 23.6 & 6.13 & 17.97 & 7.09 & 20.03 & 5.04 \\
PLLuM-70B-Inst        & 57.4 & 39.97 & 39.14 & 18.73 & 38.43 & \textbf{20.39} & \textbf{48.63} & \textbf{24.67} & 57.12 & 40.0 & 54.94 & \textbf{41.77} & 35.83 & \textbf{18.94} \\
\midrule
LLaMA3-8B             & 34.3 & 15.35 & 29.02 & 10.11 & 28.81 & 10.97 & 31.07 & 11.68 & 38.92 & 20.36 & 43.8 & 25.06 & 31.74 & 13.62 \\
Qwen3.5-9B            & 70.54 & 7.47 & 47.23 & 3.52 & 50.41 & 3.55 & 58.16 & 4.77 & 72.07 & 13.15 & 74.94 & 29.37 & 47.68 & 11.99 \\
Gemma-3-12B           & 57.68 & 2.63 & 40.37 & 1.5 & 40.29 & 2.24 & 46.41 & 1.66 & 61.26 & 5.59 & 63.04 & 41.27 & 42.23 & 17.71 \\
Gemma-3-27B           & 65.42 & 43.43 & 46.35 & 22.25 & 44.26 & 23.95 & 52.32 & 28.33 & 66.67 & 50.99 & 68.1 & 46.08 & 45.37 & 23.3 \\
Qwen3.5-35B-A3B       & 81.05 & 59.75 & 56.64 & 29.64 & 64.43 & 36.89 & 70.1 & 38.74 & 79.1 & 58.56 & 81.77 & 56.46 & 60.22 & 29.97 \\
LLaMA3.3-70B          & 73.72 & 50.48 & 52.95 & 26.47 & 54.73 & 28.43 & 60.02 & 34.3 & 74.95 & 50.63 & 77.22 & 48.61 & 51.36 & 24.25 \\
Qwen3.5-122B & \underline{\textbf{83.96}} & \underline{\textbf{62.79}} & \underline{\textbf{62.88}} & \underline{\textbf{33.16}} & \underline{\textbf{66.55}} & \underline{\textbf{39.67}} & \underline{\textbf{72.32}} & \underline{\textbf{42.43}} & \underline{\textbf{82.16}} & \underline{\textbf{60.72}} & \underline{\textbf{83.54}} & \underline{\textbf{63.04}} & \underline{\textbf{63.08}} & \underline{\textbf{37.87}} \\
\midrule
MedGemma-4B           & 39.28 & 15.49 & 30.96 & 10.82 & 30.67 & 9.66 & 31.89 & 10.35 & 44.14 & 17.84 & 42.03 & 19.49 & 31.2 & 11.17 \\
MediPhi-Instruct       & 31.81 & 9.96 & 28.41 & 9.67 & 27.39 & 8.3 & 27.64 & 7.87 & 33.33 & 12.61 & 46.33 & 26.58 & 36.78 & 15.26 \\
BioMistral-7B         & 24.34 & 9.27 & 20.23 & 6.6 & 21.67 & 8.27 & 20.14 & 4.24 & 20.72 & 11.53 & 19.24 & 13.42 & 17.98 & 7.77 \\
BioMistral-7B-DARE    & 18.95 & 12.59 & 18.56 & 8.88 & 18.66 & 9.73 & 20.82 & 6.63 & 18.56 & 15.32 & 21.52 & 15.95 & 20.3 & 8.45 \\
MedGemma-27B & \textbf{65.56} & 29.46 & \textbf{43.18} & 10.47 &\textbf{44.03} & 14.33 & \textbf{49.9} & 18.15 & \textbf{68.47} & 38.02 & \textbf{70.13} & 13.67 & \textbf{46.32} & 4.5 \\
OpenBioLLM-70B        & 54.63 & 38.73 & 36.5 & 18.65 & 41.14 & 20.97 & 47.26 & 24.74 & 62.88 & 44.68 & 61.52 & 42.78 & 38.42 & \textbf{21.66} \\
Meditron3-70B         & 53.11 & \textbf{46.06} & 32.1 & \textbf{21.99} & 37.2 & \textbf{26.19} & 43.86 & \textbf{31.59} & 57.66 & \textbf{50.09} & 50.63 & \textbf{45.57} & 35.42 & 20.71 \\
\bottomrule
\end{tabular}
}
\caption{Model evaluation results. OM = Our evaluation methodology used on original questions, MS = \textit{Multiple Statements} results on the same set of questions as OM}
\label{tab:original_vs_multistatement}
\end{table*}58.6774.8166.82Multiple Statements

The results of the evaluation on the \textbf{MS, and AS} questions (Table \ref{tab:original_vs_multistatement} and \ref{tab:tab:original_vs_multianswer}) present a significant drop in the accuracies between the original questions and the multiple response ones, showcasing the vulnerabilities in models' reasoning and their medical knowledge.
\begin{table*}
\resizebox{\textwidth}{!}{
\begin{tabular}{lcccccccccccccc}
\toprule
 & \multicolumn{14}{c}{\textbf{Exam Performance (\%)}} \\
\cmidrule(lr){2-15}
\textbf{Model} &
\multicolumn{2}{c}{LEK} &
\multicolumn{2}{c}{LDEK} &
\multicolumn{2}{c}{PES} &
\multicolumn{2}{c}{PESDL} &
\multicolumn{2}{c}{PESF} &
\multicolumn{2}{c}{LEK-EN} &
\multicolumn{2}{c}{LDEK-EN} \\
\cmidrule(lr){2-3}\cmidrule(lr){4-5}\cmidrule(lr){6-7}
\cmidrule(lr){8-9}\cmidrule(lr){10-11}\cmidrule(lr){12-13}\cmidrule(lr){14-15}
 & OM & MA & OM & MA & OM & MA & OM & MA & OM & MA & OM & MA & OM & MA \\
\midrule
Bielik-4.5B           & 46.34 & 33.33 & 43.83 & 22.22 & 43.07 & 22.67 & 46.21 & 23.79 & 41.18 & 35.29 & 49.38 & 33.33 & \textbf{50.00} & 22.62 \\
Bielik-11B            & \textbf{63.01} & \textbf{44.31} & 46.91 & \textbf{32.10} & 50.38 & \textbf{30.23} & \textbf{53.28} & \textbf{26.55} & 70.59 & \textbf{47.06} & \textbf{63.58} & \textbf{47.53} & 46.43 & \textbf{23.81} \\
PLLuM-12B-NC-Inst     & 41.06 & 25.20 & 32.72 & 24.07 & 35.26 & 18.64 & 36.72 & 17.41 & 47.06 & 29.41 & 40.74 & 14.81 & 36.90 & 13.10 \\
PLLuM-12B-NC-Chat     & 44.31 & 26.02 & 35.19 & 23.46 & 39.04 & 18.14 & 38.97 & 19.83 & 41.18 & 29.41 & 39.51 & 17.90 & 29.76 & 16.67 \\
PLLuM-12B-Inst        & 30.89 & 19.11 & 30.25 & 15.43 & 35.77 & 13.60 & 31.38 & 14.83 & 17.65 & 11.76 & 30.25 & 12.35 & 27.38 & 11.90 \\
PLLuM-12B-Chat        & 39.43 & 25.20 & 22.84 & 13.58 & 33.25 & 15.37 & 27.76 & 16.72 & 35.29 & 17.65 & 18.52 & 12.35 & 17.86 & 9.52 \\
PLLuM-70B-Inst        & 59.76 & 26.83 & \textbf{51.85} & 18.52 & \textbf{52.90} & 19.40 & 49.83 & 18.28 & \textbf{76.47} & 17.65 & 55.56 & 29.63 & \textbf{50.00} & \textbf{23.81} \\
\midrule
LLaMA3-8B             & 50.00 & 19.51 & 46.30 & 16.05 & 49.87 & 16.12 & 64.66 & 16.03 & 52.94 & 23.53 & 54.32 & 16.05 & 46.43 & 16.67 \\
Qwen3.5-9B            & 76.83 & 23.58 & 58.02 & 21.60 & 60.20 & 17.63 & 71.38 & 15.86 & 76.47 & 11.76 & 74.07 & 27.78 & \underline{\textbf{66.67}} & 23.81 \\
Gemma-3-12B           & 50.41 & 18.29 & 43.21 & 17.90 & 44.84 & 17.63 & 46.03 & 16.21 & 64.71 & 11.76 & 60.49 & 24.07 & 48.81 & 21.43 \\
Gemma-3-27B           & 61.79 & 20.33 & 50.62 & 16.05 & 54.41 & 14.61 & 60.69 & 16.03 & 52.94 & 17.65 & 63.58 & 21.60 & 53.57 & 19.05 \\
Qwen3.5-35B-A3B       & \underline{\textbf{82.52}} & 55.28 & 64.81 & \underline{\textbf{44.44}} & 70.28 & 41.81 & 77.24 & 46.03 & 82.35 & 58.82 & 80.86 & 37.65 & 58.33 & 33.33 \\
LLaMA3.3-70B          & 71.54 & \underline{\textbf{56.50}} & 61.11 & 37.65 & 62.47 & \underline{\textbf{43.32}} & 67.76 & \underline{\textbf{48.10}} & 76.47 & \underline{\textbf{70.59}} & 69.75 & \underline{\textbf{58.02}} & 59.52 & \underline{\textbf{44.05}} \\
Qwen3.5-122B-A10B     & \underline{\textbf{82.52}} & 55.28 & \underline{\textbf{70.37}} & 41.36 & \underline{\textbf{71.54}} & \underline{\textbf{43.32}} & \underline{\textbf{78.10}} & 44.66 & \underline{\textbf{82.35} }& 58.82 & \underline{\textbf{82.10}} & 51.23 & \underline{\textbf{66.67}} & 42.86 \\
\midrule
MedGemma-4B           & 47.56 & 25.61 & 48.77 & 19.14 & 45.59 & 16.62 & 51.03 & \textbf{21.55} & 29.41 & 11.76 & 49.38 & 40.12 & 46.43 & 15.48 \\
MediPhi-Instruct      & 47.15 & 18.29 & 46.30 & 16.67 & 49.37 & 17.88 & 51.21 & 13.28 & 47.06 & 29.41 & 45.06 & 28.40 & 45.24 & 14.29 \\
BioMistral-7B         & 41.06 & 13.01 & 44.44 & 12.96 & 44.33 & 9.32 & 50.86 & 5.52 & 35.29 & 11.76 & 37.04 & 17.90 & 40.48 & 11.90 \\
BioMistral-7B-DARE    & 41.87 & 11.79 & 42.59 & 12.96 & 45.09 & 10.08 & 51.55 & 6.55 & 29.41 & 5.88 & 40.12 & 13.58 & 45.24 & 9.52 \\
MedGemma-27B          & \textbf{68.29 }& \textbf{22.76} & 51.85 & 19.14 & 55.42 & 14.61 & 61.38 & 17.93 & 52.94 & 17.65 & \textbf{66.67 }& 24.69 & \textbf{58.33} & 22.62 \\
OpenBioLLM-70B        & 60.98 & 20.73 & \textbf{54.94} & 19.14 & 54.16 & \textbf{20.91} & \textbf{65.00} & 17.24 & \textbf{76.47} & 17.65 & 62.35 & 22.22 & \textbf{58.33} & 13.10 \\
Meditron3-70B         & 55.69 &\textbf{ 22.76} & 53.09 & \textbf{26.54} & \textbf{55.92} & 20.65 & 59.14 & 19.31 & \textbf{76.47} & \textbf{35.29} & 59.26 & \textbf{35.19} & 55.95 & \textbf{30.95} \\
\bottomrule
\end{tabular}
}
\caption{Model evaluation results. OM = Our evaluation methodology used on original questions, MA = \textit{Multiple Answers} results on the same set of questions as OM}
\label{tab:tab:original_vs_multianswer}
\end{table*}Multiple Answers

When it comes to the \textbf{AS} results (Table \ref{tab:corr_ans_sub_vs_original}), which present the models' ability to abstain when there is no correct answer, the observed drop in accuracy is substantial. The only model that presents improvement over the original evaluations is MedPhi, which on Polish exams obtains an accuracy of over 50\%. However, in English, it does not showcase the same trend. Since it is an English-centered model, there might be some training bias that has interfered with the evaluation method in Polish. 
\begin{table*}
\resizebox{\textwidth}{!}{
\begin{tabular}{lcccccccccccccc}
\toprule
 & \multicolumn{14}{c}{\textbf{Exam Performance (\%)}} \\
\cmidrule(lr){2-15}
\textbf{Model} &
\multicolumn{2}{c}{LEK} &
\multicolumn{2}{c}{LDEK} &
\multicolumn{2}{c}{PES} &
\multicolumn{2}{c}{PESDL} &
\multicolumn{2}{c}{PESF} &
\multicolumn{2}{c}{LEK-EN} &
\multicolumn{2}{c}{LDEK-EN} \\
\cmidrule(lr){2-3}\cmidrule(lr){4-5}\cmidrule(lr){6-7}
\cmidrule(lr){8-9}\cmidrule(lr){10-11}\cmidrule(lr){12-13}\cmidrule(lr){14-15}
 & OM & AS & OM & AS & OM & AS & OM & AS & OM & AS & OM & AS & OM & AS \\
\midrule
Bielik-4.5B           & 55.92 & 8.57 & 40.52 & 9.17 & 35.47 & 6.64 & 44.47 & 7.6 & 50.64 & 4.63 & 54.96 & 20.84 & 39.36 & 16.47 \\
Bielik-11B      & 72.52 & \textbf{21.0} & 54.14 & \textbf{15.93} & 49.32 & \textbf{17.43} & 58.88 & \textbf{18.67} & 62.47 & \textbf{23.14} & 70.42 & \textbf{16.3 }& 48.82 & \textbf{12.58} \\
PLLuM-12B-NC-Inst     & 53.0 & 6.54 & 39.17 & 5.63 & 35.06 & 6.05 & 42.69 & 5.04 & 51.16 & 6.17 & 45.8 & 5.8 & 36.4 & 6.17 \\
PLLuM-12B-NC-Chat     & 53.78 & 6.84 & 40.01 & 4.45 & 35.22 & 5.91 & 43.09 & 5.01 & 53.98 & 7.2 & 46.64 & 4.12 & 36.06 & 3.63 \\
PLLuM-12B-Inst        & 50.74 & 8.98 & 38.44 & 7.32 & 35.76 & 8.57 & 41.9 & 7.8 & 46.79 & 7.71 & 46.39 & 4.29 & 32.85 & 5.07 \\
PLLuM-12B-Chat        & 41.7 & 8.98 & 29.83 & 7.71 & 29.42 & 8.19 & 35.0 & 6.73 & 37.53 & 6.43 & 36.39 & 7.56 & 27.2 & 6.42 \\
PLLuM-70B-Inst        &\textbf{ 74.54} & 5.59 &\textbf{ 53.74} & 2.87 & \textbf{55.23} & 3.88 & \textbf{65.7} & 4.62 & \textbf{67.1} & 5.4 & \textbf{74.62} & 4.96 & \textbf{53.97} & 1.77 \\
\midrule
LLaMA3-8B             & 44.68 & 22.84 & 35.62 & 20.37 & 37.07 & 23.86 & 42.72 & 20.3 & 46.02 & 17.74 & 62.1 & 24.79 & 43.58 & 22.04 \\
Qwen3.5-9B            & 76.44 & 37.66 & 55.77 & 24.48 & 56.91 & 29.15 & 69.53 & 31.65 & 73.52 & 42.16 & 77.9 & 29.75 & 57.69 & 19.68 \\
Gemma-3-12B           & 66.69 & 9.58 & 49.86 & 5.46 & 47.56 & 7.48 & 58.8 & 7.07 & 59.9 & 9.51 & 70.0 & 11.76 & 48.48 & 7.94 \\
Gemma-3-27B           & 74.54 & 9.88 & 54.98 & 8.05 & 54.69 & 9.63 & 65.0 & 8.22 & 63.24 & 9.77 & 74.62 & 15.71 & 52.36 & 11.23 \\
Qwen3.5-35B-A3B       & 85.78 & 48.13 & 65.84 & 25.21 & 69.96 & 34.3 & 80.12 & 39.45 & 78.92 & 40.36 & 85.55 & 44.71 & 67.31 & 24.24 \\
LLaMA3.3-70B          & 78.41 & 12.61 & 60.1 & 6.53 & 60.9 & 8.7 & 71.56 & 9.49 & 73.01 & 11.83 & 82.61 & 14.12 & 59.04 & 6.84 \\
Qwen3.5-122B          & \underline{\textbf{88.28}} &\textbf{ 52.77} & \underline{\textbf{72.03}} & \textbf{33.71} & \underline{\textbf{74.24}} & \textbf{39.13} & \underline{\textbf{84.46}} & \textbf{47.79} & \underline{\textbf{80.21}} & \textbf{50.9} & \underline{\textbf{88.99}} &\underline{\textbf{ 56.97}} &\underline{\textbf{ 70.19}} &\underline{\textbf{ 40.71}} \\
\midrule
MedGemma-4B           & 50.62 & 5.41 & 39.9 & 3.83 & 38.12 & 4.18 & 45.03 & 3.83 & 49.61 & 4.37 & 58.91 & 5.21 & 40.12 & 3.97 \\
MediPhi-Instruct      & 33.49 &\underline{\textbf{ 61.04}} & 30.44 & \underline{\textbf{60.66}} & 30.21 &\underline{\textbf{ 58.57}} & 36.24 & \underline{\textbf{57.59}} & 41.65 &\underline{\textbf{ 60.67}} & 63.45 & 20.08 & 42.82 & 16.05 \\
BioMistral-7B         & 33.55 & 3.81 & 27.86 & 2.98 & 28.63 & 3.25 & 33.6 & 3.86 & 38.05 & 3.08 & 46.64 & 1.09 & 34.21 & 1.35 \\
BioMistral-7B-DARE    & 34.09 & 6.6 & 27.29 & 6.36 & 27.68 & 6.24 & 32.7 & 5.83 & 31.88 & 4.11 & 49.33 & 2.69 & 35.05 & 2.45 \\
MedGemma-27B          & 75.67 & 10.23 & 53.97 & 6.02 & 54.04 & 7.29 & \textbf{63.93} & 8.14 & 67.61 & 12.34 & 76.39 & 14.45 & 56.84 & 8.78 \\
OpenBioLLM-70B        & 70.14 & 25.52 & 51.27 & 21.05 & \textbf{54.07} & 22.26 & 62.29 & 21.85 & 65.55 & 25.71 & \textbf{76.81} & \textbf{31.43} & 55.66 & \textbf{25.42} \\
Meditron3-70B         &\textbf{ 71.56} & 33.19 & \textbf{55.94} & 18.4 & 52.96 & 26.03 & 63.48 & 25.96 & \textbf{67.87} & 24.16 & 75.88 & 10.34 & \textbf{57.18} & 5.49 \\
\bottomrule
\end{tabular}
}
\caption{Model evaluation results. OM = Our evaluation methodology used on original questions, AS = \textit{Correct answer substitution} results on the same set of questions as OM}
\label{tab:corr_ans_sub_vs_original}
\end{table*}

\begin{table*}[h]
\centering
\resizebox{0.91\textwidth}{!}{
\begin{tabular}{lcccccccccccccc}
\toprule
 & \multicolumn{14}{c}{\textbf{Exam Performance (\%)}} \\
\cmidrule(lr){2-15}
\textbf{Model} &
\multicolumn{2}{c}{LEK} &
\multicolumn{2}{c}{LDEK} &
\multicolumn{2}{c}{PES} &
\multicolumn{2}{c}{PESDL} &
\multicolumn{2}{c}{PESF} &
\multicolumn{2}{c}{LEK-ENG} &
\multicolumn{2}{c}{LDEK-ENG} \\
\cmidrule(lr){2-3}\cmidrule(lr){4-5}\cmidrule(lr){6-7}
\cmidrule(lr){8-9}\cmidrule(lr){10-11}\cmidrule(lr){12-13}\cmidrule(lr){14-15}
 & OM & PM & OM & PM & OM & PM & OM & PM & OM & PM & OM & PM & OM & PM \\
\midrule
Bielik-4.5B           & 47.97 & 43.53 & 37.36 & 33.35 & 32.15 & 30.58 & 31.21 & 34.20 & 23.92 & 27.31 & 52.68 & 50.52 & 39.55 & 37.98 \\
Bielik-11B            & \textbf{70.69} & \textbf{71.19} &\textbf{ 52.27} & \textbf{51.62} & 48.12 & \textbf{47.85} & 54.39 & \textbf{53.30} & 48.77 & \textbf{48.65} & 69.91 & \textbf{67.06 }& 46.68 & \textbf{45.34} \\
PLLuM-12B-NC-Inst     & 48.58 & 48.39 & 36.04 & 35.49 & 32.91 & 32.49 & 37.30 & 35.84 & 31.23 & 32.22 & 42.81 & 39.57 & 32.04 & 31.23 \\
PLLuM-12B-NC-Chat     & 49.30 & 48.33 & 37.04 & 35.18 & 32.94 & 32.75 & 37.59 & 36.29 & 32.92 & 31.40 & 43.80 & 43.86 & 32.95 & 33.48 \\
PLLuM-12B-Inst        & 46.83 & 36.06 & 34.99 & 27.78 & 32.65 & 25.14 & 35.20 & 27.83 & 30.58 & 22.92 & 40.75 & 28.72 & 30.16 & 24.22 \\
PLLuM-12B-Chat        & 38.90 & 21.21 & 29.16 & 17.77 & 27.49 & 14.96 & 30.66 & 14.88 & 23.92 & 12.92 & 31.32 & 24.72 & 25.22 & 19.09 \\
PLLuM-70B-Inst        & 70.39 & 45.64 & 51.46 & 32.60 & \textbf{50.43} & 31.12 & \textbf{57.67} & 35.78 & \textbf{47.84} & 27.72 & \textbf{71.56} & 60.96 & \textbf{50.38} & 41.83 \\
\midrule
LLaMA3-8B             & 44.31 & 45.90 & 36.77 & 38.55 & 35.93 & 37.41 & 40.82 & 42.58 & 31.99 & 34.09 & 60.74 & 61.22 & 42.30 & 43.21 \\
Qwen3.5-9B            & 75.85 & 74.37 & 55.42 & 52.82 & 55.80 & 53.63 & 64.30 & 62.59 & 54.80 & 52.98 & 78.48 &  2.44 & 56.85 &  1.10 \\
Gemma-3-12B           & 64.21 & 53.83 & 48.29 & 37.98 & 45.90 & 36.16 & 53.08 & 42.30 & 46.20 & 37.66 & 68.49 & 61.82 & 47.25 & 41.55 \\
Gemma-3-27B           & 72.00 & 67.75 & 53.78 & 48.67 & 52.23 & 48.00 & 59.75 & 56.81 & 50.47 & 50.64 & 74.07 & 70.10 & 52.07 & 49.00 \\
Qwen3.5-35B-A3B         & 84.67 & 84.09 & 65.45 & 63.23 & 68.73 & 66.36 & 75.06 & 72.58 & 60.23 & 58.77 & 85.12 & 85.62 & 66.21 & 66.05 \\
LLaMA3.3-70B          & 78.21 & 79.02 & 59.71 & 59.71 & 59.98 & 59.70 & 66.35 & 67.29 & 56.43 & 55.56 & 81.21 & 82.23 & 58.82 & 61.23 \\
Qwen3.5-122B-A10B     & \underline{\textbf{87.06}} & \underline{\textbf{87.71 }}& \underline{\textbf{71.00}} &\underline{\textbf{ 71.15}} & \underline{\textbf{72.19}} &\underline{\textbf{ 72.44}} &\underline{ \textbf{77.45}} &\underline{\textbf{ 78.69}} & \underline{\textbf{61.64}} & \underline{\textbf{62.98}} & \underline{\textbf{87.46}} & \underline{\textbf{87.78} }& \underline{\textbf{68.65} }& \underline{\textbf{69.68}} \\
\midrule
MedGemma-4B           & 49.66 & 43.76 & 39.28 & 34.19 & 37.86 & 32.78 & 42.45 & 36.41 & 36.08 & 29.53 & 58.24 & 53.09 & 40.05 & 36.51 \\
MediPhi-Instruct      & 33.57 & 26.73 & 31.99 & 26.27 & 29.87 & 24.61 & 34.27 & 26.62 & 26.32 & 20.64 & 60.87 & 55.70 & 42.08 & 38.11 \\
BioMistral-7B         & 32.17 & 19.63 & 28.89 & 19.51 & 27.35 & 18.52 & 29.83 & 19.16 & 21.64 & 12.57 & 42.24 & 32.53 & 32.45 & 25.59 \\
BioMistral-7B-DARE    & 31.68 & 23.67 & 28.62 & 23.09 & 26.33 & 21.47 & 30.05 & 22.78 & 20.00 & 17.08 & 45.10 & 34.72 & 33.45 & 27.53 \\
MedGemma-27B          & \textbf{73.54} & \textbf{70.39} & \textbf{53.07} & \textbf{49.70} &\textbf{ 52.47} & \textbf{49.77 }& \textbf{58.92} & \textbf{55.57} & \textbf{51.35} & \textbf{50.88} & \textbf{76.39} & 73.53 &\textbf{ 54.29 }& 48.50 \\
OpenBioLLM-70B        & 67.45 & 59.24 & 49.51 & 41.76 & 50.62 & 40.84 & 57.43 & 47.56 & 48.83 & 35.03 & 73.85 & \textbf{74.67} & 52.97 & \textbf{52.41} \\
Meditron3-70B         & 65.69 & 25.53 & 49.82 & 32.26 & 48.44 & 21.55 & 54.44 & 19.42 & 47.08 & 15.15 & 71.88 &  2.60 & 53.07 &  1.22 \\
\bottomrule
\end{tabular}
}
\caption{Model evaluation results expressed as percentages. \textit{OM} = Our evaluation Methodology, \textit{PM} = Previous evaluation Methodology.}
\label{tab:short_bias_nobias}
\end{table*}

\begin{table*}
\centering
\resizebox{0.9\textwidth}{!}{
\begin{tabular}{lccccccc}
\toprule
 & \multicolumn{7}{c}{\textbf{Original vs modified avg difference (\%)}} \\
\cmidrule(lr){2-8}
\makecell{\textbf{Model}} & LEK & LDEK & PES & PESDL & PESF & LEK-ENG & LDEK-ENG \\
\midrule
Bielik-4.5B & -32.56 & -23.64 & -21.73 & \textbf{-17.79} & \textbf{-12.07} & -30.90 & -24.39 \\
Bielik-11B & -46.13 & -35.65 & -30.32 & -37.77 & -40.33 & -47.51 & -33.77 \\
PLLuM-12B-NC-Inst & -34.02 & -24.61 & -22.20 & -26.22 & -17.28 & -32.21 & -23.67 \\
PLLuM-12B-NC-Chat & -35.77 & -27.67 & -23.18 & -25.95 & -20.71 & -35.09 & -25.70 \\
PLLuM-12B-Inst & -32.48 & -25.26 & -22.64 & -26.49 & -24.04 & -32.91 & -22.92 \\
PLLuM-12B-Chat & \textbf{-27.81} & \textbf{-20.85} & \textbf{-20.26} & -24.37 & -23.00 & \underline{\textbf{-22.55}} & \underline{\textbf{-17.92}} \\
PLLuM-70B-Inst & -51.57 & -38.73 & -37.37 & -42.89 & -35.90 & -52.47 & -38.04 \\
\midrule
LLaMA3-8B & \textbf{-21.65} & \textbf{-17.42} & \textbf{-16.23} &\textbf{ -23.24 }& \textbf{-22.69} & -33.08 &\textbf{ -20.75} \\
Qwen3.5-9B & -46.76 & -36.15 & -36.05 & -45.91 & -47.87 & -47.39 & -37.37 \\
Gemma-3-12B & -54.30 & -41.49 & -38.53 & -46.99 & -53.49 & -47.94 & -34.14 \\
Gemma-3-27B & -50.81 & -38.01 & -35.15 & -41.84 & -31.33 & -49.16 & -33.97 \\
Qwen3.5-35B & -32.23 & -34.53 & -32.08 & -35.94 & -27.88 & -37.55 & -37.62 \\
LLaMA3.3-70B & -49.35 & -42.28 & -40.19 & -43.17 & -38.91 & -54.04 & -41.59 \\
Qwen3.5-122B & -30.83 & -34.65 & -31.51 & -33.53 & -24.66 & \textbf{-29.31} & -27.67 \\
\midrule
MedGemma-4B & -37.32 & -29.96 & -28.64 & -31.89 & -33.82 & -42.33 & -29.82 \\
MediPhi-Instruct &\underline{\textbf{ 9.03}} &\underline{\textbf{ 9.11}} & \underline{\textbf{6.40}} &\underline{\textbf{ -0.95}} &\underline{\textbf{ -4.58}} & -35.33 & -24.98 \\
BioMistral-7B & -25.62 & -21.08 & -21.31 & -25.10 & -19.87 & \textbf{-34.17} & \textbf{-24.56} \\
BioMistral-7B-DARE & -21.99 & -17.25 & -17.40 & -22.93 & -13.53 & -35.17 & -25.33 \\
MedGemma-27B & -55.69 & -41.83 & -39.78 & -44.56 & -40.59 & -58.73 & -45.29 \\
OpenBioLLM-70B & -36.29 & -26.09 & -27.38 & -33.40 & -27.68 & -38.50 & -25.93 \\
Meditron3-70B & -29.25 & -27.03 & -21.25 & -26.95 & -22.79 & -47.87 & -37.15 \\
\bottomrule
\end{tabular}
}
\caption{Weighted (by the number of questions in each question group) average differences between the original question scores and the modified (Multiple Statement -- MS, Multiple Answer -- MA, Correct answer substitution -- AS).}
\label{tab:avg_delat}
\end{table*}

\clearpage

\section{Instructions}
\label{sec:instructions}
\begin{table}[h!]
\centering
\begin{tabular}{p{15cm}}
\toprule
Your task is to provide answers to a medical test for doctors. From all the provided answers A, B, C, D, E select only one. If you are not sure, choose the most probable one. Answer in a manner: Correct answer is B\\
\midrule
\caption{Instruction for inferring the answer used in \textbf{Previous Metodology (PM)}, with a sample answer.}
\end{tabular}

\end{table}

\begin{table}[h!]
\centering
\begin{tabular}{p{15cm}}
\toprule
Your task is to provide answers to a medical test for doctors. From the provided answers, select the correct one and respond only with that answer. End your answer with a period.\\
\midrule
\caption{Instruction for inferring the answer used in \textbf{Our Metodology (OM)}, with a sample answer.}
\end{tabular}

\end{table}
\begin{table}[h!]
\centering
\begin{tabular}{p{15cm}}
\toprule
Your task is to provide answers to a medical test for doctors. From the provided statements, select all that are correct and return only their indexes. End your answer with a period.\\
\midrule
\caption{Instruction for inferring the answer for a \textbf{Multiple Statements} type of question.}
\end{tabular}

\end{table}

\begin{table}[h!]
\centering
\begin{tabular}{p{15cm}}
\toprule
Your task is to provide answers to a medical test for doctors. From the provided options, select all the correct ones and respond only with those answers. End your answer with a period.\\
\midrule\caption{Instruction for inferring the answer for a \textbf{Multiple Answers} type of question.}
\end{tabular}
\end{table}

\begin{table*}[h!]
\centering
\begin{tabular}{p{15cm}}
\toprule
Your task is to provide a precise and specialized answer to a medical question for doctors. Answer with one sentence, phrase, medical term, most appropriate clinical course of action, or diagnosis.\\
\midrule
\caption{Instruction for inferring the answer for an \textbf{Open-Ended} type of question.}
\end{tabular}
\end{table*}

\begin{table*}[h!]
\centering

\begin{tabular}{p{15cm}}
\toprule
Take on the role of a medical examiner in an oral exam. You ask the doctor a question (QUESTION), and the doctor provides a proposed answer (PROPOSED ANSWER). You also know the correct answer (ORIGINAL ANSWER). Based on your objective judgment, medical knowledge, and known guidelines, is the PROPOSED ANSWER correct? Do you determine that the PROPOSED ANSWER is the correct answer to the QUESTION? If the PROPOSED ANSWER is correct return True, if the PROPOSED ANSWER is incorrect return False. Do not return anything more than a Boolean. \\
\midrule
\caption{Instruction used in \textbf{LLM-as-a-judge}.}
\end{tabular}
\end{table*}

\clearpage

\clearpage

\section{Question Modification Examples}
\label{sec:modification-examples}

\subsection{Multiple Statements (MS)}

\textbf{Original question}

\begin{tcolorbox}[questionbox]
Indicate true statements regarding complications associated with using chemotherapy in cancer treatment:

\begin{enumerate}[label=\arabic*)]
    \item the most frequent haematological complication is neutropenia (found in 60--88\% of the patients treated);
    \item neutropenic fever is found in ca.\ 10--50\% of patients treated for solid tumours and in over 80\% of patients treated for haematological malignancies;
    \item the G-CSF prophylaxis is recommended only in radical and palliative treatment.
\end{enumerate}

The correct answer is:

\begin{enumerate}[label=\Alph*.]
    \item 1,2
    \item all of the above
    \item 1,3
    \item 2 only
    \item 3 only
\end{enumerate}
\end{tcolorbox}

\textbf{Correct answer:} \textbf{A}

\vspace{1em}

\textbf{Modified question}

\begin{tcolorbox}[questionbox]
Indicate true statements regarding complications associated with using chemotherapy in cancer treatment:

\begin{enumerate}[label=\arabic*)]
    \item the most frequent haematological complication is neutropenia (found in 60--88\% of the patients treated);
    \item neutropenic fever is found in ca.\ 10--50\% of patients treated for solid tumours and in over 80\% of patients treated for haematological malignancies;
    \item the G-CSF prophylaxis is recommended only in radical and palliative treatment.
\end{enumerate}
\end{tcolorbox}

\textbf{Correct answer:} \textbf{1, 2}

\newpage
\subsection{Multiple Answers (MA)}

\textbf{Original question}

\begin{tcolorbox}[questionbox]
Which of the following activities are characteristic of six-month-old infants?

\begin{enumerate}[label=\Alph*.]
    \item supporting their body on extended arms with partly or fully opened hands
    \item bringing a toy from one hand to the other
    \item dropping things on purpose
    \item answers A and B are correct
    \item answers A, B, and C are correct
\end{enumerate}
\end{tcolorbox}

\textbf{Correct answer:} \textbf{D}

\vspace{1em}

\textbf{Modified question}

\begin{tcolorbox}[questionbox]
Which of the following activities are characteristic of six-month-old infants?

\begin{enumerate}[label=\Alph*.]
    \item supporting their body on extended arms with partly or fully opened hands
    \item bringing a toy from one hand to the other
    \item dropping things on purpose
\end{enumerate}
\end{tcolorbox}

\textbf{Correct answer:} \textbf{A, B}

\newpage

\subsection{Correct Answer Substitution (AS)}
\label{app:correct_ans_sub}

\textbf{Original question}

\begin{tcolorbox}[questionbox]
An elderly male patient with obturative lung disease was diagnosed with hernia. It was protruding from the abdominal cavity through the transverse fascia forming the posterior wall of the inguinal canal, bordered superiorly by the conjoint tendon, inferiorly by the inguinal ligament, and laterally by the inferior epigastric vessels.

The hernia in such location is known as:

\begin{enumerate}[label=\Alph*.]
    \item oblique inguinal hernia
    \item scrotal hernia
    \item direct inguinal hernia
    \item femoral hernia
    \item spigelian hernia
\end{enumerate}
\end{tcolorbox}

\textbf{Correct answer:} \textbf{C}

\vspace{1em}

\textbf{Modified question}

\begin{tcolorbox}[questionbox]
An elderly male patient with obturative lung disease was diagnosed with hernia. It was protruding from the abdominal cavity through the transverse fascia forming the posterior wall of the inguinal canal, bordered superiorly by the conjoint tendon, inferiorly by the inguinal ligament, and laterally by the inferior epigastric vessels.

The hernia in such location is known as:

\begin{enumerate}[label=\Alph*.]
    \item oblique inguinal hernia
    \item scrotal hernia
    \item none of the answers is correct
    \item femoral hernia
    \item spigelian hernia
\end{enumerate}
\end{tcolorbox}

\textbf{Correct answer:} \textbf{C}

\newpage
\subsection{Open-Ended (OE)}

\textbf{Original question}

\begin{tcolorbox}[questionbox]
A 36-year-old multiparous woman went to see a gynecologist because of regular but excessive and painful periods which has lasted for the last few years. This is accompanied by increasing fatigue, general weakness, and more frequent urination. The pathological findings included pale mucosa. Gynecological exam revealed a tumour the size of a 4-month pregnancy. 
Blood test and transvaginal ultrasound were made. What lab test results and diagnosis can you expect?

\begin{enumerate}[label=\Alph*.]
    \item anaemia, uterine myomas
    \item anaemia, pregnancy
    \item normal blood count, pregnancy
    \item normal blood count, simple ovarian cyst
    \item anaemia, simple ovarian cyst
\end{enumerate}
\end{tcolorbox}

\textbf{Correct answer:} \textbf{A}

\vspace{1em}

\textbf{Modified question}

\begin{tcolorbox}[questionbox]
A 36-year-old multiparous woman went to see a gynecologist because of regular but excessive and painful periods which has lasted for the last few years. This is accompanied by increasing fatigue, general weakness, and more frequent urination. The pathological findings included pale mucosa. Gynecological exam revealed a tumour the size of a 4-month pregnancy. 
Blood test and transvaginal ultrasound were made. What lab test results and diagnosis can you expect?
\end{tcolorbox}

\textbf{Correct answer:} \textbf{anaemia, uterine myomas}

\newpage
\section{Contamination Studies}
\label{contamination}
\subsection{DCQ -- Data Contamination Quiz}
\begin{figure*}[t]
    \centering
    \includegraphics[width=\linewidth]{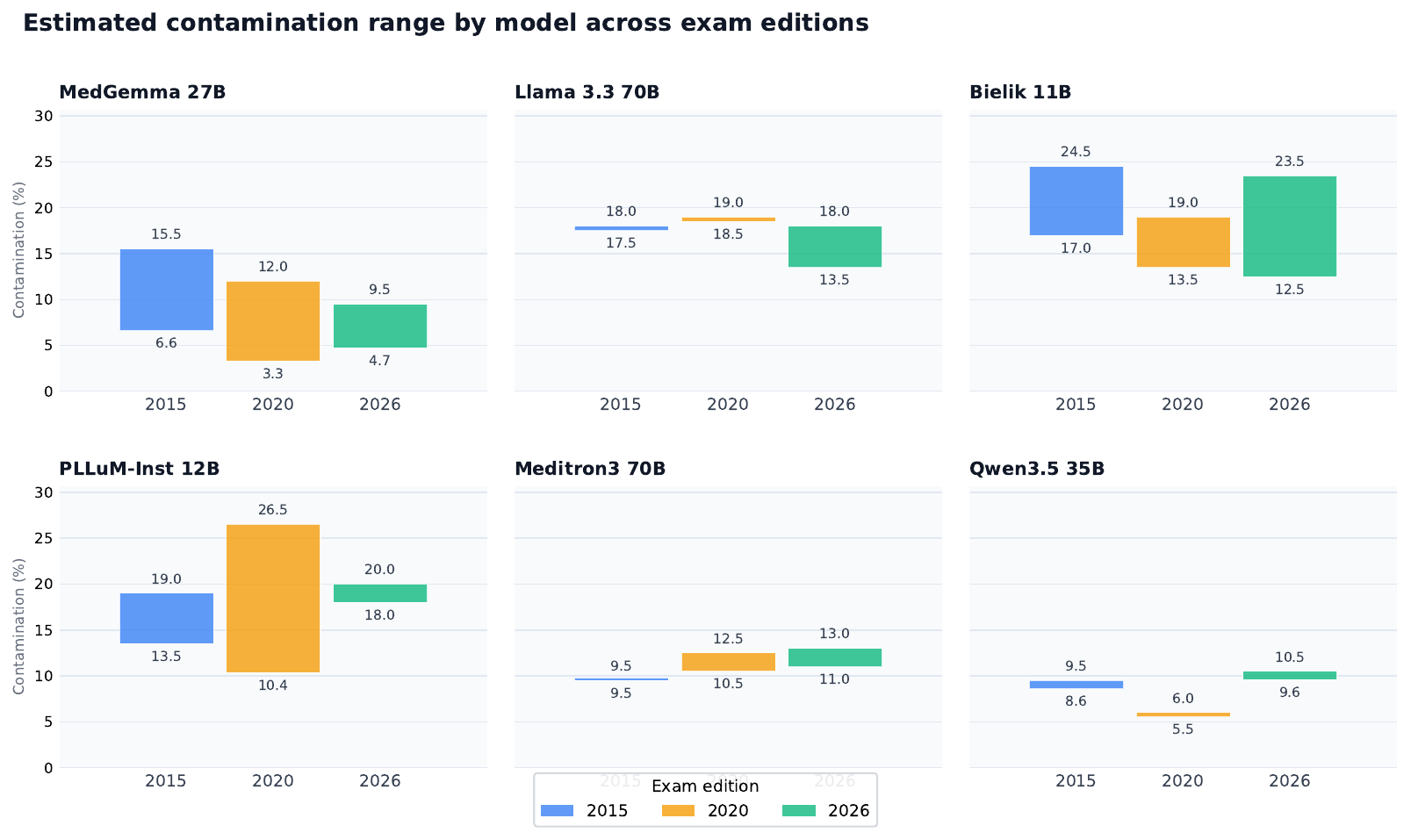}
    \caption{Estimated contamination in LLMs from different groups and model families, across three spring editions 2015, 2020, 2026 of combined LEK and LDEK exams.}
    \label{fig:cont_lek_ldek}
\end{figure*}

DCQ is a lightweight method introduced by \citet{DCQ} for evaluating data contamination in a model’s training data. 
The method is based on creating a quiz in which the model must identify the original instances from a target dataset among several perturbed variants of those instances. 
We created such a quiz for LEK and LDEK exam questions from three different spring editions (2015, 2020, and 2026). This setup allows us to assess whether the editions suspected of being leaked into the training data (2015 and 2020) exhibit higher contamination scores than the edition that could not have been leaked (2026). The contamination estimations across the editions (shown in Figure \ref{fig:cont_lek_ldek}) do suggest limited data leakage into the training data. However, the estimated contamination for the 2026 edition is not significantly lower than that for 2015 and 2020, as we observed. It is worth pointing out that in the DCQ we are measuring the contamination of the exam questions only, which are far more available than the exam answers.

\subsection{Web Search Audit of Question and Answer Availability}

Polish examination questions are publicly available, but access is restricted by a CAPTCHA system. To check whether they may be included in LLM training corpora, we audited whether they could be found through a web search. For this experiment, we used the 2023 autumn editions of LEK and PES and queried Tavily\footnote{\url{https://www.tavily.com/}} for each question without the answer. We then compared the best retrieved text fragment with the original question using normalized Levenshtein distance.

\begin{figure*}[t]
\centering
\includegraphics[width=\linewidth]{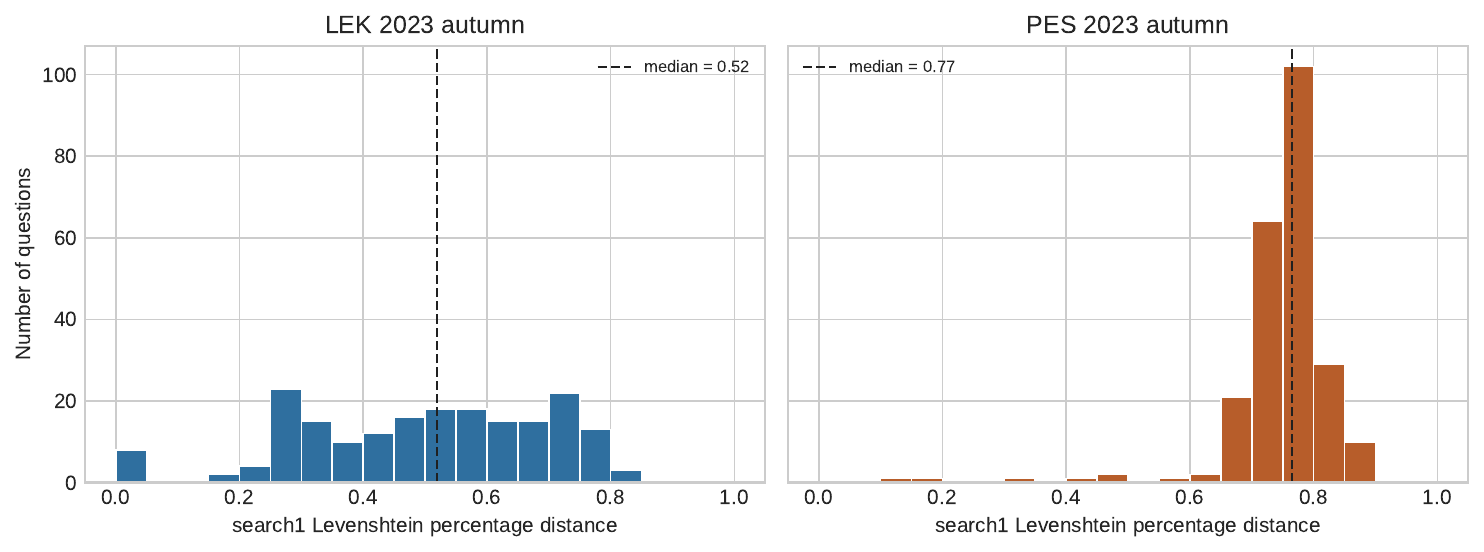}
\caption{Distribution of normalized Levenshtein distances between the original question and the closest text fragment retrieved by Tavily for LEK and PES. Lower values indicate closer textual matches.}
\label{fig:tavily_search1_levenshtein_hist}
\end{figure*}

Figure~\ref{fig:tavily_search1_levenshtein_hist} shows the distribution of normalized character-level Levenshtein distances for LEK and PES. The distributions differ substantially. LEK questions were found much more frequently in near-exact or highly similar form on the web, while PES matches were generally more distant. The median distance was 0.52 for LEK and 0.77 for PES.

We then performed a second, stricter analysis for LEK. Instead of relying on the short search preview, we retrieved the full content of the returned page and asked \texttt{gpt-5-mini} to judge whether the page contained the same question and whether it also contained the correct answer. We treated a case as evidence of potential answer leakage only when both conditions were satisfied.

For LEK, \texttt{gpt-5-mini} judged 169 of 194 returned pages to contain the same or a materially equivalent question. However, most of these pages did not expose the answer. This pattern was consistent with many results pointing to paid LEK learning portals, where the question text is visible but the answer is hidden behind access control. In total, 36 of 194 pages contained the correct answer. Some of these answer-containing cases came from freely available flashcard-style learning platforms.

For PES, we did not find evidence of data leakage in which both the same question and its corresponding answer were available together in the top returned web result. This is consistent with the character of the exam: PES is a highly specialized board-certification examination, less commonly targeted by public learning materials. By contrast, LEK is taken by every candidate seeking the right to practice medicine in Poland, and the larger ecosystem of preparation materials makes repeated or highly similar questions easier to find online. The observed pattern is also consistent with candidates' perception of the exams: LEK preparation materials are widely available, whereas PES materials are substantially scarcer. Some LEK leakage likely reflects the fact that questions repeat, or appear in very similar form, across consecutive years.

\subsection{Temporal Performance Analysis}

Additionally, to check whether the LLMs perform better on older exams (which would prove data leakage),  we conducted a temporal contamination analysis using the English versions of LEK and LDEK. We focused on the 2024 spring editions, whose questions were released after the training cutoff of the majority of the evaluated models and, therefore, could not have been directly memorized during pre-training.

\begin{figure*}[t]
    \centering
    \includegraphics[width=\linewidth]{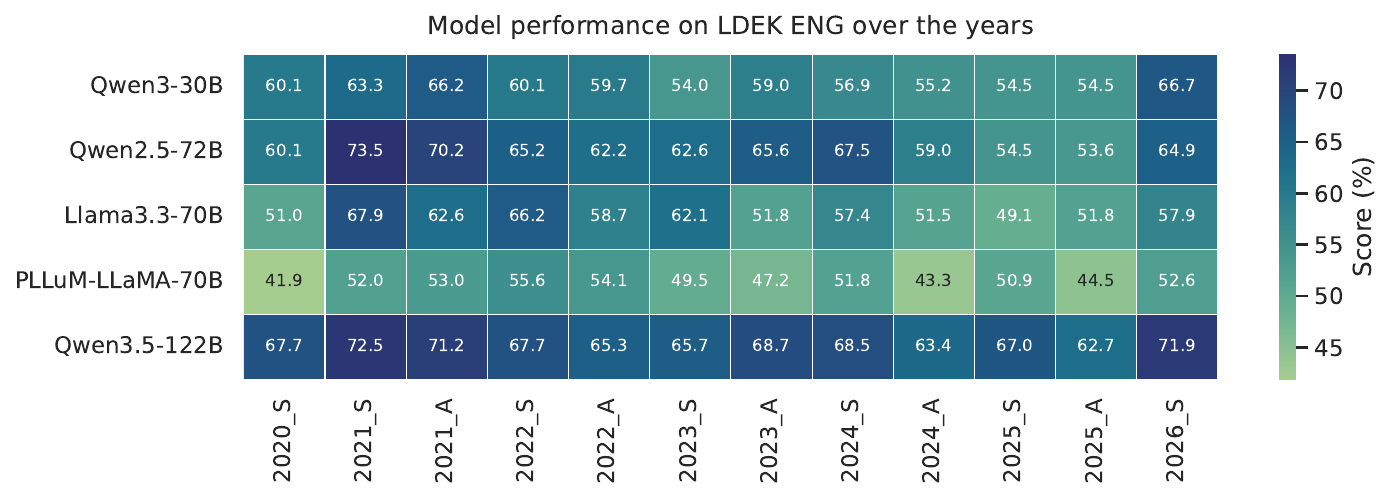}
    \caption{Model performance on LDEK in English over the editions Spring-\textbf{S} Autumn-\textbf{A} over the years}
    \label{fig:year_ldek}
\end{figure*}

\begin{figure*}[t]
    \centering
    \includegraphics[width=\linewidth]{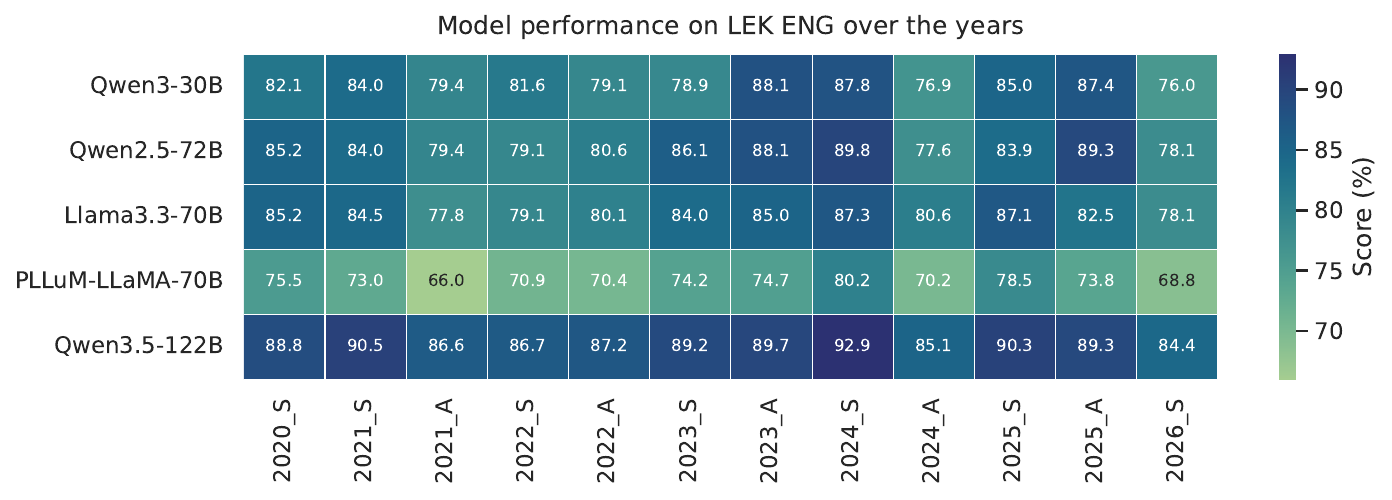}
    \caption{Model performance on LEK in English over the editions Spring-\textbf{S} Autumn-\textbf{A} over the years}
    \label{fig:year_lek}
\end{figure*}

If contamination were the primary explanation for high benchmark performance, one would expect a systematic degradation across recent editions relative to earlier years, depending on the model's knowledge cutoff date. However, such a pattern was not consistently observed. In several cases, performance on the 2025 data was comparable to the average performance on examinations from 2020--2024, whereas performance in 2026 was not a significant outlier. No visible deterioration in performance over recent years is observed in the LEK ENG and LDEK ENG exams, as presented in Figures \ref{fig:year_ldek} and \ref{fig:year_lek}.

\section{Regular Expressions for Question Filtering}

We use regular expressions to (i) identify parts of questions that require removal or modification and (ii) filter questions eligible for a given modification group (e.g., MA-group).

For the MA-group, we apply the following language-specific patterns:

\textbf{Polish:}
\begin{lstlisting}
pattern = r"prawdziwe sa odpowiedzi"
filtered = [p if re.search(pattern, p, re.IGNORECASE) else "" for p in parts]
pattern = r"wszystkie\s+.*?\s+(prawidlowe|prawdziwe)"
filtered2 = [p for p in parts if re.search(pattern, p, re.IGNORECASE)]
\end{lstlisting}

\textbf{English:}
\begin{lstlisting}
pattern = r"(?:are\s+(?:correct|true)|true\s+(?:answers|statements)|correct\s+answers?|answers|\&)"
filtered = [p if re.search(pattern, p, re.IGNORECASE) else "" for p in parts]
\end{lstlisting}

\section{Detailed Summary of Disagreement Analysis}
\label{sec:disagreement_analysis}

After calculating agreement metrics for human annotators and AI judges in the evaluation of open-ended answers, we performed an additional analysis of cases in which their assessments diverged. The aim was to identify recurring sources of disagreement, define them as reproducible discrepancy patterns, and examine how these patterns were treated by human annotators and LLM-as-a-judge systems.

\subsection{Analytical Process}
The analysis consisted of three stages: preliminary review of evaluated cases, supplementary annotation of discrepancy patterns, and pattern-level analysis of evaluator behaviour.

\subsubsection{Preliminary case review}
A senior annotator who was also a physician reviewed 15 control cases completed by all six human annotators and 74 disagreement cases in which a single human annotator’s decision contradicted the unanimous assessment of seven AI judges (gpt-5.4-nano, gpt-5.4-mini, gpt-5.5, claude-haiku-4-5, claude-sonnet-4-6, claude-opus-4-7, gpt-oss-120b). For each case, the senior annotator reviewed all materials available to, and produced by, the initial annotator, including the original question, the official correct answer, the AI-generated answer, the human annotator’s decision, and, when available, their written commentary.

The source of disagreement was not always identifiable, and in some cases appeared unique to a single answer. However, multiple outputs shared recurring features, and their relevance was further supported by the available initial annotators’ comments.

The main outcome of this preliminary review was the definition of six \textbf{discrepancy patterns}: recurring features of evaluated answers that were repeatedly suspected of contributing to disagreement between human and AI evaluators, and in some cases also to variability among human annotators. These patterns form a practical framework for describing the main sources of disagreement in the analysed sample, rather than a complete taxonomy of answer defects.

The discrepancy patterns used in the supplementary annotation were defined as follows:
\begin{itemize}
    \item \textbf{A – Defects in redundant content:} The output addressed the issue indicated by the official correct answer but also included additional content that introduced at least one decision-relevant defect.
    \item \textbf{B – Valid alternative interpretation of the question:} The output answered a valid interpretation of the open-ended question, but not the issue implied by the original closed-ended question and its official correct answer.
    \item \textbf{C – Underspecified expected detail:} The output addressed the expected issue but did so with lower specificity or completeness than the official correct answer.
    \item \textbf{D – Non-existent medical terms:} The output contained a term absent from Polish, English, or Latin medical nomenclature, including invalid hybrid terms.
    \item \textbf{E – Misuse of existing medical terms:} The output used an existing Polish, English, or Latin medical term inconsistently with its accepted meaning.
    \item \textbf{F – Incorrect spelling or notation of medical terms:} The output contained an existing Polish, English, or Latin medical term with incorrect spelling, notation, inflection, abbreviation, or written form.
\end{itemize}

The detailed criteria used for applying these labels are presented in Table~\ref{tab:discrepancy_patterns_criteria}.
\begin{table}
    \centering
    \resizebox{\linewidth}{!}{%
    \begin{tabular}{cll}
        \toprule
        Discrepancy pattern& Human evaluator tendency & AI evaluator tendency \\
        \midrule
        A & Penalised errors appearing in both necessary and redundant content.& Accepted answers if expected information was present, disregarding errors in redundant content.\\
        B & Accepted valid alternative interpretations if they contained no significant errors.& Usually rejected answers that missed information specified in answer key.\\
        C & No consistent tendency identified. & No consistent tendency identified. \\
        D & Usually penalised non-existent medical terms. & Usually did not penalise non-existent medical terms. \\
        E & Too few cases to infer a dominant tendency. & Too few cases to infer a dominant tendency. \\
        F & Usually penalised incorrect spelling or notation. & Usually did not penalise incorrect spelling or notation. \\
        \bottomrule
    \end{tabular}
    }
    \caption{Observed tendencies in evaluator approach to discrepancy patterns. See Table~\ref{tab:discrepancy_patterns_overview} for code (A-F) definitions.}
    \label{tab:evaluator_approaches}
\end{table}

\begin{table*}
    \centering
    \begin{tabular}{p{0.05\textwidth}p{0.25\textwidth}p{0.6\textwidth}}
        \hline
        \multicolumn{2}{c}{\textbf{Discrepancy pattern}}& Criteria\\
        \hline
        A & Defects in redundant content & 1) Output engages with the issue indicated by the official correct answer.
\textit{and}
2) Output includes additional content beyond that issue.
\textit{and}
3) The additional content features $\ge$ 1 defect that could influence the evaluator’s decision.\\
\midrule
        B & Valid alternative interpretation of the question & 1) Output addresses a valid interpretation of the open-ended question.
\textit{and}
2) This interpretation differs from the issue expressed by the official correct answer.\\
\midrule
        C & Underspecified expected detail & 1) Output addresses the expected issue expressed by the official correct answer.
\textit{and}
2) Output addresses this issue with lower specificity or completeness than the official correct answer.\\
\midrule
        D & Non-existent medical terms & 1) Output contains a term absent from Polish, English, or Latin medical nomenclature.
\textit{or}
2) Output contains a non-standard hybrid medical term defined as an expression created by combining Polish, English, or Latin elements into a form that is not itself recognised in medical nomenclature.\\
\midrule
        E & Misuse of existing medical terms & 1) Output contains an existing Polish, English or Latin medical term.
\textit{and}
2) Output uses this term inconsistently with its accepted meaning.\\
\midrule
        F & Incorrect spelling or notation of medical terms & 1) Output contains an existing Polish, English or Latin medical term.
\textit{and}
2) Spelling, notation, inflection or abbreviation of this term is incorrect.\\
        \bottomrule
    \end{tabular}
        
    \caption{Criteria used for applying discrepancy patterns to LLM answers.}
    \label{tab:discrepancy_patterns_criteria}
\end{table*}

\subsubsection{Supplementary annotation}
Of the 74 discrepant cases, 3 were excluded because the official correct answer was inconsistent with current medical knowledge. The senior annotator then applied the six discrepancy patterns to the remaining 71 cases to describe potential sources of disagreement.

Each answer could receive all applicable labels, provided that their criteria were compatible. Patterns A, B, and C were mutually exclusive, as they described different relations between the output and the issue implied by the original closed-ended question and its official correct answer. Patterns D–F could co-occur with any compatible A–C pattern and with one another; for example, an answer containing a non-existent medical term in an unnecessary explanation could be labelled as both D and A.

\subsubsection{Pattern-level analysis of evaluator behaviour}
After supplementary annotation, we analysed the prevalence of each discrepancy pattern and the distribution of positive and negative human evaluations among answers assigned that pattern. These results served as supporting evidence for the potential influence of each characteristic on human decisions.

For reference, we also prepared an outlier-excluded summary. The excluded annotator’s 28 annotations had been rejected because their performance on control questions differed from that of the remaining human annotators. This summary served as a supplementary sensitivity analysis, assessing whether the observed pattern distribution and evaluation outcomes remained broadly similar after removing cases linked to atypical control-case performance.

Tables ~\ref{tab:discrepancy_patterns_w_outlier} and ~\ref{tab:discrepancy_patterns_wo_outlier} present the prevalence of each discrepancy pattern and the distribution of positive and negative human assessments among answers assigned that pattern, with Table~\ref{tab:discrepancy_patterns_w_outlier} including all assessments and Table~\ref{tab:discrepancy_patterns_wo_outlier} excluding assessments made by the outlier annotator.

\begin{table*}[h]
    \centering
    \resizebox{\textwidth}{!}{
    \begin{tabular}{lllll}
        \toprule
        \multicolumn{2}{c}{Discrepancy pattern}& Answers assigned pattern, n (\%)& Positive human assessments, n (\%)& Negative human assessments, n (\%)\\
        \midrule
        A & Defects in redundant content & 14 (20\%) & 0 (0\%) & 14 (100\%) \\
        B & Valid alternative interpretation of the question & 20 (28\%) & 19 (95\%) & 1 (5\%) \\
        C & Underspecified expected detail & 12 (17\%) & 8 (67\%) & 4 (33\%) \\
        D & Non-existent medical terms & 14 (20\%) & 2 (14\%) & 12 (86\%) \\
        E & Misuse of existing medical terms & 4 (6\%) & 1 (25\%) & 3 (75\%) \\
        F & Incorrect spelling or notation of medical terms & 10 (14\%) & 1 (10\%) & 9 (90\%) \\
        \bottomrule
    \end{tabular}
    }
    \caption{Prevalence of discrepancy patterns and distribution of human assessments among answers assigned each pattern, including assessments made by the outlier annotator}
    \textit{Note.} Percentages for pattern prevalence use all 71 included answers as the denominator; percentages for positive and negative assessments use answers assigned the given pattern. Patterns were not mutually exclusive.
    \label{tab:discrepancy_patterns_w_outlier}
\end{table*}

\begin{table*}[h]
    \centering
    \resizebox{\textwidth}{!}{
    \begin{tabular}{lllll}
        \toprule
        \multicolumn{2}{c}{Discrepancy pattern}& Answers assigned pattern, n (\%)& Positive human assessments, n (\%)& Negative human assessments, n (\%)\\
        \midrule
        A & Defects in redundant content & 14 (33\%) & 0 (0\%) & 14 (100\%) \\
        B & Valid alternative interpretation of the question & 7 (16\%) & 7 (100\%) & 0 (0\%) \\
        C & Underspecified expected detail & 7 (16\%) & 3 (43\%) & 4 (57\%) \\
        D & Non-existent medical terms & 11 (26\%) & 1 (9\%) & 10 (91\%) \\
        E & Misuse of existing medical terms & 2 (5\%) & 0 (0\%) & 2 (100\%) \\
        F & Incorrect spelling or notation of medical terms & 7 (16\%) & 0 (0\%) & 7 (100\%) \\
        \bottomrule
    \end{tabular}
    }
    \caption{Prevalence of discrepancy patterns and distribution of human assessments among answers assigned each pattern, excluding assessments made by the outlier annotator}
    \textit{Note.} Percentages for pattern prevalence use all 43 included answers as the denominator; percentages for positive and negative assessments use answers assigned the given pattern. Patterns were not mutually exclusive.
    \label{tab:discrepancy_patterns_wo_outlier}
\end{table*}

\paragraph{Limitations}
These data should be interpreted cautiously. A single AI-generated answer could contain more than one discrepancy pattern or other defects that were not separately coded. Therefore, the final annotator decision could not be attributed to one pattern alone. Some answers were counted in more than one group. For example, an answer containing a non-existent term in redundant content was counted under both A and D. Similarly, an answer containing a real medical term that was both misused and misspelled was counted under both E and F.

Conclusions about the dominant approaches of human and AI evaluators to these patterns were based not only on pattern frequency and evaluation outcomes, but also on the initial annotators’ comments and the senior annotator’s qualitative case-by-case observations.

\subsection{Findings}

\subsubsection{Pattern-specific findings}
Below we present our findings regarding each discrepancy pattern. 
Where relevant, we also indicate whether a given pattern appeared to reflect limitations of the original closed-to-open question conversion, the behaviour of the tested LLMs, or differences in how human and AI evaluators approached cases characterised by these features. All data reported in this subsection refer to the analysis, including cases annotated by the outlier annotator

\paragraph{A – Defects in redundant content}
This pattern was identified in 14 AI-generated answers. In two cases, the only defect was an incorrect expression used alongside a correct term as if it were an equivalent. In the remaining cases, defects appeared in unnecessary explanations, elaborations, or specifications.

We interpret this pattern mainly as a consequence of the tested LLMs’ tendency to expand answers beyond the issues explicitly requested in the question. However, none of the relevant questions explicitly prohibited such expansion.

Human annotators tended to reject answers containing errors they considered important, regardless of whether the error appeared in the core part of the answer or in redundant content. By contrast, LLM-as-a-judge systems tended to accept answers if they contained the expected information, even when other parts of the answer included significant defects.

\paragraph{B – Valid alternative interpretation of the question}
This pattern was identified in 20 AI-generated answers. In most cases, it appeared to result from loss of context during conversion of closed-ended questions into open-ended form. Without the original answer options, the question could often be answered in more than one reasonable way.

Only one case could be interpreted as a genuine misunderstanding of the question. In the remaining cases, the LLM answer was valid for the open-ended wording of the question but did not match the interpretation implied by the original answer key. This included three questions asking the respondent to identify an incorrect statement. With the answer options removed, the LLMs generated their own examples of false statements.

Human annotators tended to accept answers if they were appropriate to the wording of the open-ended question and did not contain important errors. LLM-as-a-judge systems tended to reject outputs when they did not match the answer key.

\paragraph{C – Underspecified expected detail}
This pattern was identified in 12 answers. In each case, the question did not explicitly define the expected level of detail, such as the level of biological organisation or number of elements required.

We interpret this pattern as another consequence of context loss during conversion from closed-ended to open-ended questions. In the original format, the answer options expressed a predefined level of specificity. After conversion, this information was no longer available.

Evaluator approaches to this pattern were heterogeneous. Some less specific answers were accepted, whereas others were rejected. Variation was also observed between individual human annotators. No clear dominant approach could be identified among either human annotators or LLM-as-a-judge systems.

\paragraph{D – Non-existent medical terms}
This pattern was identified in 14 answers. Human annotators rejected most answers affected by this pattern, and often treated such defects as important because they directly concerned medical nomenclature and could indicate lack of terminological reliability. In contrast, LLM-as-a-judge systems usually did not appear to penalise such defects, particularly when the intended meaning could be inferred or when the answer otherwise matched the expected content.

\paragraph{E – Misuse of existing medical terms}
This pattern was identified in 4 answers. Because the number of cases was very small, the data were considered insufficient to infer a dominant evaluation tendency among human annotators or AI judges.

This pattern nevertheless remained conceptually important because it captured a distinct type of terminological defect: the use of real medical terms in a meaning inconsistent with accepted medical usage.

\paragraph{F – Incorrect spelling or notation of medical terms}
This pattern was identified in 10 answers. It was usually associated with negative human evaluations and was often explicitly mentioned in annotator comments as a reason for rejection.

LLM-as-a-judge systems did not appear to penalise this pattern in most cases. They often accepted answers despite incorrect spelling or notation of medical terms, especially when the intended term was recognisable and the broader answer overlapped with the expected content.

\subsubsection{Effects of Outlier Exclusion on Pattern Distribution}
The outlier-excluded summary generally supported the main interpretation of evaluator behaviour, although it also showed that some estimates of pattern prevalence were sensitive to the excluded annotator’s cases. In both summaries, defects in redundant content, non-existent medical terms, and incorrect spelling or notation were predominantly associated with negative human assessments, whereas valid alternative interpretations of the question were predominantly accepted. The largest change concerned the prevalence of valid alternative interpretations, which decreased after excluding the outlier annotator, indicating that this pattern was overrepresented among that annotator’s cases. The mixed distribution of assessments for underspecified expected detail further supported the absence of a consistent evaluation approach for this pattern.

\subsection{Interpretation and Implications}

Overall, the disagreement analysis indicates that discrepancies between human annotators and AI judges were driven mainly by systematic differences in how they handled context loss, redundant content, and terminology-related defects, while some variability was also observed among human annotators. Human annotators were generally more sensitive to clinically or linguistically important defects, whereas LLM-as-a-judge systems tended to prioritise overlap with the answer key and often overlooked errors outside the expected content.

These findings suggest that at least some of the observed differences may be mitigable through more explicit evaluation instructions for both human annotators and LLM-as-a-judge systems, specifically targeting the recurring sources of discrepancy identified in this analysis.

\end{document}